\documentclass[Afour,sageh,times]{sagej}

\makeatletter
\renewcommand{\thesection}{\arabic{section}}          

\renewcommand{\@seccntformat}[1]{{\csname the#1\endcsname}\hspace{0.5em}}
\makeatother

\usepackage{moreverb,url}
\usepackage[table]{xcolor} 

\definecolor{myblue}{RGB}{42, 73, 161}
\definecolor{myred}{RGB}{212, 45, 42}

\usepackage{hyperref} 

\hypersetup{
    colorlinks = true,   
    pdfborder = {0 0 0}, 
    linkcolor  = myred,
    citecolor  = myblue,
    urlcolor   = magenta
}

\usepackage[T1]{fontenc}
\usepackage[utf8]{inputenc}
\usepackage{amsmath,amsfonts}
\usepackage{algorithmic}
\usepackage[ruled,linesnumbered]{algorithm2e}
\usepackage{array}
\usepackage{bm}
\usepackage[caption=false,font=normalsize,labelfont=sf,textfont=sf]{subfig}
\usepackage{textcomp}
\usepackage{stfloats}
\usepackage{verbatim}
\usepackage{graphicx}
\usepackage{cite}
\usepackage{tabularx,booktabs}
\usepackage{multirow}
\usepackage{multicol}
\usepackage{colortbl} 
\usepackage{xcolor}
\usepackage{hyperref}
\usepackage{amssymb}
\usepackage{bbding}
\usepackage{enumitem}
\usepackage{appendix}  

\usepackage{siunitx}

\DeclareMathOperator{\diag}{diag}

\newcommand{\bb}{\boldsymbol}

\renewcommand{\epsilon}{\varepsilon}
\renewcommand{\phi}{\varphi}

\newcolumntype{L}[1]{>{\raggedright\arraybackslash}p{#1}}
\newcolumntype{C}[1]{>{\centering\arraybackslash}p{#1}}
\newcolumntype{R}[1]{>{\raggedleft\arraybackslash}p{#1}}

\makeatletter
\newcommand{\removelatexerror}{\let\@latex@error\@gobble}
\renewcommand{\maketag@@@}[1]{\hbox{\m@th\normalsize\normalfont#1}}%
\makeatother
\usepackage{makeidx}

\newcommand\BibTeX{{\rmfamily B\kern-.05em \textsc{i\kern-.025em b}\kern-.08em
T\kern-.1667em\lower.7ex\hbox{E}\kern-.125emX}}

\def\volumeyear{2026}

\begin{document}


\title{\texttt{ExoTraj}: A General Lower-limb Exoskeleton Assistance Policy for Complex Environments}

\author{Xiao-Yin Liu\affilnum{1,2}, Guotao Li\affilnum{1,2}, Long Sun\affilnum{4}, Xu Liang\affilnum{4} and Zeng-Guang Hou\affilnum{1,2,3}}

\affiliation{\affilnum{1}The State Key Laboratory of Multimodal Artificial Intelligence Systems, Institute of Automation, Chinese Academy of Sciences, China\\
\affilnum{2}The School of Artificial Intelligence, University of Chinese Academy of Sciences, China\\
\affilnum{3}CASIA-MUST Joint Laboratory of Intelligence Science and Technology, Institute of Systems Engineering, Macau University of Science and Technology, China.\\
\affilnum{4}The School of Automation and Intelligence, Beijing
Jiaotong University, China}

\corrauth{Guotao Li and Zeng-Guang Hou}

\email{guotao.li@ia.ac.cn and zengguang.hou@ia.ac.cn}

\begin{abstract}
    Adaptive torque prediction in dynamic exoskeleton scenarios requires expensive motion capture systems, which are infeasible in complex outdoor environments.
    Trajectory prediction has emerged as one of the effective approaches to address such an issue. 
    However, the core challenges of exoskeleton trajectory prediction are twofold: 1) establishing the mapping from multi-modal features to trajectory information; 2) constructing the mapping from trajectory to torque.
    For the former, most existing methods perform only single-step prediction in dynamic scenarios and neglect inter-subject trajectory variability, thereby limiting the trajectory optimization space and prediction generalization.
    To address this, this paper proposes a fast flow matching (FM) method that enables accurate trajectory prediction and better generalization with accelerated inference for real-time performance, where trajectory generation errors and encoded observations are used to guide the training direction. 
    For the second challenge, due to the high dynamics of the human-robot system and the strong coupling between perception and control, simple control methods struggle to achieve efficient assistance based on the predicted trajectory. 
    This paper utilizes model predictive control (MPC) and designs a novel optimization objective to optimize torque, ensuring the exoskeleton achieves comfortable and robust assistance.
    By integrating the above two components, the unified policy, denoted as \texttt{ExoTraj}, is developed to enable adaptive assistance in complex outdoor scenarios without high data acquisition cost. 
    Experimental results show that compared to traditional methods, \texttt{ExoTraj} reduces cross-subject prediction error by $14.0\%$ during the online phase and maintains robustness against external noise. Relative to the zero torque condition, \texttt{ExoTraj} decreases metabolic rate by $11.5\%\thicksim24.4\%$, heart rate by $1.7\%\thicksim19.5\%$, and peak muscle activation levels by $10.9\%\thicksim41.3\%$, respectively. This progress facilitates the deployment of exoskeletons toward viable community‑wide adoption. Project page: \href{https://xiaoyinliu0714.github.io/Home_ExoTraj/}{ExoTraj}.
\end{abstract}


\maketitle
	\section{Introduction}
With the global aging population continuously expanding, the number of individuals with impaired walking abilities and stroke-related conditions is steadily increasing. Exoskeleton robots have emerged as a promising solution to support elderly care \citep{slade2022personalizing}. When the assistance provided by exoskeletons aligns with the user’s movement intention, it can effectively enhance walking capacity in older adults and improve rehabilitation outcomes for stroke patients. Although exoskeleton robots have been deployed in relatively structured and controlled clinical environments, their application remains challenging in complex and dynamic outdoor scenarios \citep{molinaro2024estimating}. Therefore, recent research has focused on deploying exoskeletons in complex real-world environments.

\subsection{Key Challenges for Complex Environments} 
To enhance adaptability in complex environments, data-driven methods have been widely adopted for exoskeleton control. Currently, data-driven approaches can be broadly categorized into two main types: reinforcement learning (RL) based methods \citep{luo2024experiment,luo2023robust,2025TAI} and multi-modal information fusion-based methods \citep{molinaro2024task,molinaro2024estimating,divekar2024versatile}. Reinforcement learning-based methods train exoskeleton control policies through online interactions in simulated environments before transferring them to physical systems \citep{luo2023robust}. These approaches heavily rely on high-fidelity environmental simulations and accurate musculoskeletal models \citep{luo2024experiment}. However, training high-dimensional musculoskeletal models to achieve human-like walking across diverse terrains remains a highly challenging problem, which limits the potential of RL in exoskeleton applications.

On the other hand, multimodal information fusion-based methods integrate user motion state and exoskeleton state information to predict human movement intentions, thereby enabling accurate assistance \citep{molinaro2024estimating,molinaro2024task}. This class of methods holds promise for exoskeleton assistance in complex task environments.
The aforementioned method focuses on torque prediction, which relies on expensive motion capture systems (foot pressure plates are costly). However, deploying these systems in complex outdoor scenarios is impractical. To enhance the adaptability of exoskeletons in outdoor environments, it is necessary to develop a low-cost data acquisition scheme and a corresponding control framework.
In particular, human walking trajectory information, such as angle or angular velocity, is readily available in complex outdoor environments. Therefore, this paper aims to adopt a hierarchical control framework: i) the high-level controller predicts human joint trajectories (mapping from multi-modal features to trajectory); ii) the low-level controller optimizes motor torque output to track the desired trajectory (mapping from trajectory to torque). 

For the high-level controller (trajectory prediction in complex environments), existing research has predominantly focused on single-step prediction \citep{11049997,xiongmech}, which suffers from several pronounced limitations: the short prediction horizon is insufficient to compensate for the mechanical system’s response delay \citep{yi2021continuous}; the limited information cannot support trajectory-based optimization control, such as model predictive control \citep{xu2025robust} or impedance control \citep{chen2024upper}; and abrupt prediction changes may lead to discontinuous or uncomfortable assistance. Furthermore, due to the complexity of high-dimensional motion trajectories and the variability in motion patterns across users, supervised learning methods, such as LSTM and TCN, struggle to fully capture the variability in multi-modal data, often leading to low trajectory prediction accuracy and poor generalization.

For the low-level controller (torque optimization), due to the strong coupling between sensing and control of direct-drive exoskeletons, simple control methods struggle to achieve robust and comfortable control. The underlying reasons are as follows: Human locomotion is highly dynamic, and the human-exoskeleton system is a tightly coupled dynamical system \citep{molinaro2024task}. Pronounced mismatches between the predicted trajectories and the actual trajectories give rise to significant torque oscillations in the vicinity of the zero-crossing point, which in turn triggers severe anteroposterior jitter of the exoskeleton. This structural vibration further deteriorates the trajectory prediction performance of the onboard perception network, forming a detrimental positive feedback loop (vicious cycle) that directly impairs the efficacy of the exoskeleton assistance \citep{xiampc}. Thus, integrating the above high-level and low-level control, two challenges for the exoskeleton to achieve efficient, trajectory-prediction-based assistance in outdoor scenarios can be summarized as follows:
\begin{enumerate}[label=\arabic*)]
    \item Given the high-dimensional output space and user-dependent motion variability, how to effectively fuse multi-modal data from diverse users for accurate and real-time trajectory prediction and better generalization?
    \item Given the high dynamicity of human-robot systems and strong perception-control coupling, how can the optimization algorithm be designed for comfortable and robust exoskeleton control under adaptive dynamic trajectories?
\end{enumerate}

\begin{figure*}
	\centering
	\includegraphics[width=1.0\textwidth]{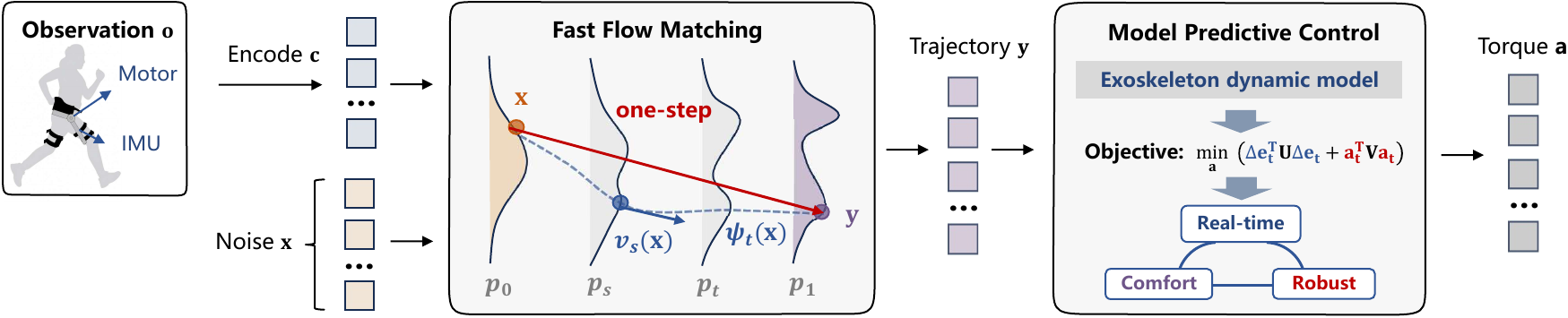}
    \captionsetup{justification=justified, singlelinecheck=false}
	\caption{The framework of the proposed assistance policy \texttt{ExoTraj}. The process begins by encoding the observation $\mathbf{o}$ to guide the learning of the velocity field $v_t$ of flow matching, and the one-step iteration enables the rapid generation of joint trajectories $\mathbf{y}$. Subsequently, a model predictive control scheme is employed to optimize the motor torque output $\mathbf{a}$ for efficient assistance. Ultimately, this pipeline ensures the delivery of safe and comfortable assistance for complex outdoor environments. }. 
	\label{fig1}
\end{figure*}

\subsection{Solutions for Key Challenges} 
For \textit{challenge 1}, human trajectory prediction is rooted in imitation learning from human locomotion. Behavior cloning, which only learns the mean trajectory of demonstrations, has limited generalization to individuals with distinct motion distributions, whereas generative methods model the full trajectory distribution and achieve superior performance in high-dimensional outputs and generalization to unseen scenarios \citep{chi2023diffusion}.
Generative methods such as diffusion model \citep{songdenoising} and flow matching (FM) \citep{lipman2022flow} have been widely exploited in applications, including robotic manipulation \citep{ding2025,chi2023diffusion} and autonomous driving \citep{fu2025moflow}. Despite their advantages in generalization, generative approaches have had very limited application in exoskeleton trajectory prediction. Compared with the diffusion model, FM provides a simpler training, as it operates by solving an ordinary differential equation rather than the more complex stochastic differential equation \citep{ding2025}. To address \textit{challenge 1}, this paper introduces flow matching into exoskeleton trajectory prediction to realize high-level fusion of multi-modal data across different subjects and improve trajectory prediction performance in complex and dynamic environments.

The high dynamicity of human locomotion in outdoor scenarios demands strong real-time performance for trajectory inference. However, FM inherently trains a vector field to solve the underlying ordinary differential equation (ODE), which requires multiple iterative steps to generate the final target and thus incurs excessive inference latency and computational overhead \citep{geng2025mean,fransone}. 
To accelerate inference, this paper develops a one-step iterative trajectory generation method that maintains high prediction accuracy while reducing inference latency.
Specifically, the error between the trajectory generated in a one-step iteration and the ground-truth trajectory is incorporated as a constraint during FM vector field training. This guides the model to generate trajectories that more closely approximate the true trajectories. Then, human-exoskeleton multi-modal information is encoded into a conditional vector to guide vector field training, enabling target locomotion trajectory generation under specific observations.
This approach can ensure both real-time performance and high accuracy in trajectory prediction under complex environments. 

For \textit{challenge 2}, the core issue lies in the high dynamics of predicted trajectories and the mutual coupling between perception and control. In dynamic scenarios, most existing works that employ predicted trajectories for exoskeleton assistance use key parameters, such as the amplitude of sinusoidal functions \citep{lee2024ai}, dynamic primitive parameters \citep{chen2024learning}, and gait parameters \citep{tricomi2025leveraging}. Due to the low variability of these trajectories, previous control methods can achieve satisfactory performance; however, they struggle to adapt to tasks with high dynamic characteristics. 
For adaptive trajectories in dynamic scenarios, the control optimization algorithm must track the desired trajectory while ensuring smooth, comfortable joint torques that align with the user’s motion intention. Model predictive control (MPC) employs receding horizon optimization, which can achieve the aforementioned objective by designing an appropriate cost function.

However, the human-robot coupled dynamic model is difficult to accurately describe in mathematical terms, especially the interaction torque. Most existing studies approximate the human-exoskeleton interaction torque via the impedance model \citep{chen2024upper}, but impedance control shares an inherent structural similarity to proportional-integral-derivative (PID) control, and is prone to inducing optimization instability when deployed in outdoor scenarios relying on high-dynamic predicted reference trajectories. Inspired by biomechanical principles, this paper employs joint angular velocity and angular acceleration information to approximate human-exoskeleton interaction torque, so as to guarantee the stability of the optimization process. To ensure the robustness and smoothness of the optimized torque profile, inspired by work \citep{2025Diff}, this paper introduces a dedicated torque penalty term in the MPC optimization cost function to constrain the output torque, thus ensuring the comfort of exoskeleton assistance.

\subsection{Results and Contributions}
Finally, combining the above two core components, a general lower-limb exoskeleton assistance policy, named \texttt{ExoTraj}, is proposed for complex outdoor environments, with its overall workflow illustrated in Fig. \ref{fig1}. \texttt{ExoTraj} adopts a hierarchical control framework: the high-level controller fuses multi-modal information to achieve accurate and real-time trajectory prediction, while the low-level controller performs real-time torque optimization to realize robust and comfortable control of the exoskeleton. We validate the performance of the proposed policy \texttt{ExoTraj} in complex dynamic outdoor scenarios, with evaluation metrics including trajectory prediction accuracy and generalization, control comfort and robustness, as well as metabolic rate, heart rate, and muscle activation.

We acquire multi-modal locomotion data, including measurements from $3$ inertial measurement units (IMUs) and $2$ drive motors, to train the high-level controller. Experimental results demonstrate that, compared with conventional data-driven methods, the improved FM method achieves a $14.0\%$ reduction in trajectory prediction error in cross-subject scenarios. MPC-optimized torque trajectory achieves robust and comfortable assistance under different types of external disturbance. Compared to the zero torque condition, the user's metabolic rate, heart rate, and peak muscle activation level are reduced by $11.5\%\thicksim24.4\%$, $1.7\%\thicksim19.5\%$, and $10.9\%\thicksim41.3\%$, respectively. The policy \texttt{ExoTraj} achieves a trajectory inference time of approximately $5$ ms, and the MPC-based torque optimization takes around $2$ ms, enabling reliable operation in highly dynamic scenarios.
The main contributions of this work can be summarized as follows:
\begin{enumerate}[label=\arabic*)]
    \item  A general assistance policy framework, named \texttt{ExoTraj}, is proposed for the hip exoskeleton, which can be adapted to diverse terrain and tasks without high data acquisition costs.
    \item A novel FM is firstly proposed for exoskeleton trajectory prediction, with one-step generation error and observation encoding guiding training, to achieve real-time, accurate prediction, and superior generalization.
    \item A new MPC optimization objective is designed for torque optimization, ensuring robust and comfortable control, and enabling real-time adaptive exoskeleton control in dynamic scenarios.
\end{enumerate}

The framework of this paper is as follows: Section \ref{related} covers related works in exoskeleton trajectory generation and control. The problem formulation for exoskeleton assistance is defined in Section \ref{problem}. The proposed trajectory generation algorithm, fast FM, is detailed in Section \ref{fm}, followed by the torque optimization and exoskeleton control framework in Section \ref{control}. Section \ref{setup} and Section \ref{Experiments} are dedicated to the experimental procedure and results.  The paper concludes this work in the final section.
    \section{Related Works}\label{related}
This section aims to systematically summarize the research work on exoskeleton trajectory prediction and assistive torque generation. Then, we elaborate on the fundamental differences between the proposed \texttt{ExoTraj} framework and existing methods reported in the literature.

\subsection{Trajectory Prediction for Exoskeleton}
Exoskeleton trajectory prediction refers to the task of predicting future multi-step trajectory information in a single forward pass based on historical observed measurements.
Traditional artificial intelligence algorithms, such as LSTM and TCN, often struggle with predicting high-dimensional temporal data (trajectory prediction). To reduce the complexity of the trajectory prediction, existing research on exoskeleton trajectory prediction mostly relies on prior analytical models (e.g., dynamic movement primitives \citep{chen2024learning} and sinusoidal functions \citep{tricomi2025leveraging}), where only the key model parameters are estimated. However, such methods suffer from poor adaptability to dynamically complex environments.

The above methods typically employ a two-stage approach for trajectory prediction. First, in the offline phase, dimensionality reduction techniques such as dynamic movement primitives (DMPs) \citep{chen2024learning} or probabilistic movement primitives (ProMPs) \citep{dong2025,chen2024upper} are applied to captured motion trajectories to learn key parameters for different types of assistance. Then, in the online phase, multi-modal sensor data are fused to classify the required assistance type, and human-in-the-loop optimization algorithms are used to refine the key parameters \citep{chen2024upper}, thereby generating the desired motion trajectory. Although the methods above can achieve satisfactory performance in scenarios with low dynamics, they often struggle in highly complex and dynamic environments due to their reliance on learning key parameters tailored only to specific static conditions.

Furthermore, there exists inherent inter-subject variability in locomotion patterns across different users, as well as intra-subject gait variability when the same individual performs identical locomotion tasks \citep{molinaro2024task}. Most existing works on lower-limb exoskeleton trajectory prediction have rarely accounted for these variabilities, which may result in limited generalization capability of the trained models. Currently, generative models are capable of efficiently fusing multimodal data and data capturing inherent variabilities, enabling high-fidelity generation of target data \citep{songdenoising,chi2023diffusion}.
Therefore, this paper aims to leverage generative models to efficiently fuse multimodal data from diverse subjects and realize real-time, high-precision trajectory prediction in complex, unstructured environments.

\subsection{Torque Generation for Exoskeleton Control}
To ensure satisfactory assistive comfort during human-exoskeleton interaction, the exoskeleton’s assistive control should use a torque-control mode. While joint torque information can be acquired via a motion capture system \citep{molinaro2024estimating}, such an approach suffers from prohibitive cost and is infeasible for deployment in complex, unstructured outdoor environments. With accessible joint torque measurements, the neural network can be trained to predict joint torque directly, without the intermediate step of trajectory forecasting \citep{molinaro2024task}. Accurate ground-truth joint torque measurements are difficult to obtain in unstructured outdoor scenarios. For this reason, most existing approaches rely on joint trajectory prediction (i.e., joint angle and angular velocity) to indirectly generate the required joint torque commands.

Existing trajectory-to-torque generation approaches for exoskeletons fall into two primary categories: optimization-based control (e.g., model predictive control \citep{xu2025robust,xiampc}) and analytical formula-based control (e.g., PID \citep{tricomi2025leveraging} and impedance control \citep{lee2024ai}). The first category of approaches consists of optimization-based methods for assistive torque determination. While model predictive control (MPC) has been widely applied to the control of lower-limb exoskeleton robots, the reference trajectories adopted in most existing works are predefined via heuristic rules and have not been implemented in closed-loop dynamic trajectory prediction pipelines. This limitation poses significant challenges to the robustness and optimization performance of MPC in complex unstructured environments.

For the second category of approaches, they exhibit limited adaptive capability in trajectory prediction, including typical methods based on sine wave peak prediction \citep{tricomi2025leveraging}, and gait phase division and terrain classification \citep{lee2024ai}. While these methods can achieve satisfactory trajectory tracking performance via PID or impedance control, they struggle to adapt to trajectory prediction tasks with high complexity and pronounced dynamic characteristics.  \citep{xiongmech} computes the torque based on the angular error between the multi-step-ahead predicted angles and the actual measured values. However, the robustness of such a method is difficult to reliably guarantee. Therefore, the aforementioned two categories of approaches are unable to reliably achieve efficient and robust assistance in complex unstructured environments. To address it, this paper develops a novel optimization objective and a dedicated estimation method for human-exoskeleton interaction torque, to realize efficient and adaptive assistive performance for the lower-limb exoskeleton.

\subsection{Comparisons and Advantages of \texttt{ExoTraj}}
This part compares the policy \texttt{ExoTraj} proposed in this paper with state-of-the-art research methods in the field (\citep{molinaro2024estimating,molinaro2024task,luo2024experiment,lee2024ai,scherpereel2025deep}), clarifies their core differences, and elaborates on the advantages of \texttt{ExoTraj} in practical deployment. Table \ref{tab_comparsion} conducts a comparison of five core dimensions: whether it relies on a motion capture system, whether it depends on a musculoskeletal model, data collection cost, training complexity, and the terrains or tasks to which it is applicable. \citet{molinaro2024estimating} and \citet{molinaro2024task} rely on motion capture systems (MoCaps) and OpenSim to acquire a large corpus of human joint torque data during motor tasks, then train a neural network to achieve one-step-ahead joint torque prediction, enabling high-efficiency assistive control for complex task scenarios. While the adopted network features a relatively straightforward training pipeline, it suffers from prohibitively high data acquisition costs and cannot be readily generalized to complex, unstructured outdoor environments. 
Building on the aforementioned MoCaps-based data, \citet{scherpereel2025deep} collects unlabeled human motion data to achieve inter-subject generalization and effectively reduces the cost of data acquisition. While this method features a significantly lowered data acquisition burden, its inherent reliance on the foundational MoCap-derived reference data constrains the integration and utilization of complementary multi-sensor modalities.

To circumvent the prohibitive cost of joint torque data acquisition, \citet{luo2024experiment} trains exoskeleton control policies in physics-based simulation environments. However, it relies on high-fidelity musculoskeletal models and elaborate reward function tuning, incurring substantial training overhead. While this method achieves high-efficiency assistive control across locomotion modes, including walking at varying speeds, running, as well as stair ascent and descent, it fails to generalize to assistive control for other complex unstructured terrains. This limitation arises as training a unified high-dimensional musculoskeletal model adaptive to complex terrains remains a long-standing open challenge. 
\citet{lee2024ai} eliminates the need for MoCaps. By fusing kinematic information from IMUs and other onboard sensors, it achieves gait pattern and terrain type recognition, and generates reference joint torques matched to the identified gait and terrain via an impedance model. While this approach features low overhead in data acquisition and model training, it is only suitable for relatively static scenarios such as walking on flat ground and stairs, and exhibits limited adaptability to highly dynamic outdoor environments.

\begin{table*}
	\normalsize
	\caption{Comparison results against state-of-the-art methods for adaptive lower-limb exoskeleton assistance in outdoor scenarios, where MoCaps denotes motion capture systems and MSK denotes musculoskeletal model.}
	\label{tab_comparsion}
	\centering
	\resizebox{17.5cm}{!}{
		\begin{tabular}{c|cc|cc|cc}
\toprule[1pt] 
\textbf{References} & \textbf{MoCaps} & \textbf{MSK} & \textbf{Data Cost} & \textbf{Complexity} & \textbf{Tasks \& Terrains} & \textbf{Scenario}\\

\midrule
\citet{molinaro2024estimating}& \Checkmark & \XSolidBrush & High &Low & Level ground, Stair and Ramp& Indoor\\
\citet{molinaro2024task}&\Checkmark& \XSolidBrush  & High &Low& Walk, Run, Jump, Squat, Step up (28 activities)& Indoor\\
\citet{scherpereel2025deep} &\Checkmark & \XSolidBrush  & Low &Low &Walk, Run, Jump, Squat, Step up (28 activities) &Indoor\\
\citet{luo2024experiment}&\XSolidBrush& \Checkmark & Low &High &Ground walk, Stair climb and Run & Outdoor\\
\citet{lee2024ai}&\XSolidBrush& \XSolidBrush& Low &Low &Level ground, Stair and Ramp& Outdoor\\
\midrule
\multirow{2}{*}{\textbf{Ours} (\texttt{ExoTraj})}&\multirow{2}{*}{\XSolidBrush}& \multirow{2}{*}{\XSolidBrush}& \multirow{2}{*}{Low} &\multirow{2}{*}{Low} &Run, Rocky climb, Rugged terrain variable-speed walk&\multirow{2}{*}{Outdoor}\\
&&&&&Jump, Squat, Kick (Multiple high-dynamic movements)\\
\bottomrule[1pt]
\end{tabular}
}
\end{table*} 

The proposed method \texttt{ExoTraj} removes the reliance on the dedicated MoCaps. With the inherent advantages of low data acquisition cost and minimal training computational complexity, it enables robust, high-performance assistive operation in complex and highly dynamic unstructured outdoor environments. In addition to the inherent advantage of low data acquisition cost, \texttt{ExoTraj} presents the following superiorities over existing methods:
\begin{enumerate}[label=\arabic*)]
    \item \textbf{Excellent cross-platform transferability:} It only requires collecting kinematic data from users wearing other exoskeletons with different sensor configurations, and can achieve assistance for the target-specific exoskeleton via rapid network retraining.
    \item \textbf{Broad scenario adaptability:} Benefiting from the portability of its data acquisition scheme, it enables high-efficiency assistive control in complex outdoor terrains and extreme environments, including rugged terrain and sloped rocky surfaces.
    \item \textbf{Superior control robustness:} By adopting a hierarchical control framework and optimizing joint torques via MPC, it achieves high-efficiency assistive control even in the presence of network prediction deviations and external disturbances.
\end{enumerate}

\section{Task Definition and Problem Formulation}\label{problem}
This section defines the task requirements for the exoskeleton robot implemented in this study and presents the key scientific problems that the paper aims to address.

\subsection{Task Definition}
Robotic exoskeleton systems represent highly coupled human–robot systems. To enhance the assistance performance of exoskeletons in complex and dynamic outdoor environments, exoskeletons must learn to predict human motion intentions and subsequently provide the corresponding assistive actions. To ensure the comfort of human–robot interaction during assistance, it is indispensable to implement a torque-controlled actuation mode for the driving motors. However, deploying expensive motion capture systems outdoors to measure the joint torques associated with human movements is impractical.
In contrast, in outdoor environments, the joint trajectories of the user can be directly acquired. Therefore, to reduce data acquisition costs and improve adaptability to complex terrains, this paper aims to predict joint trajectory information and then optimize torque output based on predicted trajectories. 

The proposed method primarily accomplishes two key tasks: first, it learns the mapping from multi-modal data (denoted as observations $\mathbf{o}_t$) to joint trajectory information $\mathbf{y}_t$; second, it establishes the mapping from joint trajectories $\mathbf{y}_t$ to torque outputs $\mathbf{a}_t$. The mathematical relationships are described as follows:
\begin{equation}\label{map}
    \begin{aligned}
        \mathbf{y}_t=\mathcal{F}(\mathbf{o}_t),\quad \quad \mathbf{a}_t=\mathcal{G}(\mathbf{y}_t),
    \end{aligned}
\end{equation}
where $\mathcal{F}(\cdot)$ and $\mathcal{G}(\cdot)$ denote the mapping functions from observations $\mathbf{o}$ to joint trajectories $\mathbf{y}$ and from joint trajectories $\mathbf{y}$ to torque outputs $\mathbf{a}$, respectively. 
The observation $\mathbf{o}_t=\{\bb{o}_{t-i}\}_{i=1}^{T_o}\in \mathbb{R}^{T_o\times H_o}$ represents the history of the state $\bb{o}$ over the past $T_o$ time steps, which consists of measurements from three IMUs and two motors, totaling $H_o$ features. The trajectory $\mathbf{y}_t=\{\bb{y}_{t+i};\bb{\dot{y}}_{t+i}\}_{i=1}^{T_p}\in \mathbb{R}^{T_p\times 2H_p}$ denotes the sequence of joint angles $\bb{y}$ and angular velocities $\bb{\dot{y}}$ over the future $T_p$-step prediction horizon, where $H_p$ is the number of joints. The torque $\mathbf{a}_t=\{\mathbf{u}_{t+i}\}_{i=0}^{N-1}\in \mathbb{R}^{N\times H_p}$ corresponds to the sequence of motor torques $\mathbf{u}$, where $N$ is the optimization horizon.

This paper aims to learn the aforementioned two mapping functions, enabling accurate real-time prediction of user motion trajectories and real-time optimization of torque outputs in complex outdoor environments, thereby ensuring the safety and compliance of the assistance. The former requires efficient fusion of multi-modal data from diverse users, while the latter necessitates the formulation of an appropriate optimization objective function. The detailed problem formulation is presented as follows.


\begin{figure*}
	\centering
	\includegraphics[width=1.0\textwidth]{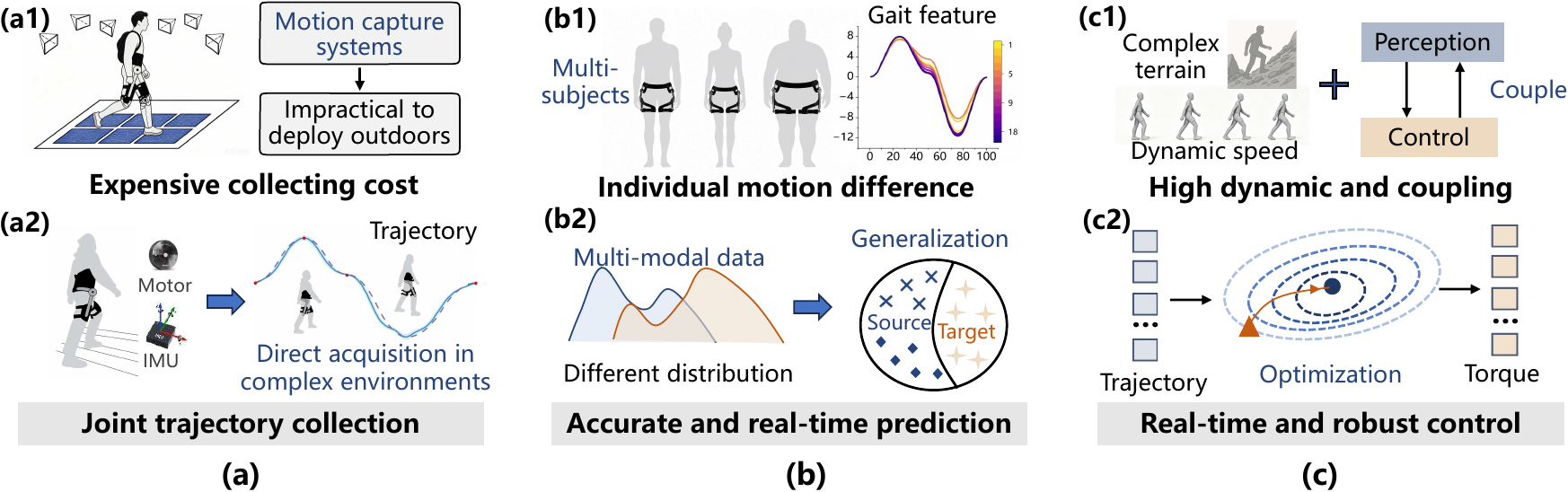}
    \captionsetup{justification=justified, singlelinecheck=false}
	\caption{Schematic overview of problems and solutions for exoskeleton assistance in complex environments. (a) shows that outdoor motion-capture-based data acquisition is cost-prohibitive and infeasible, whereas direct trajectory measurements are readily obtainable.
(b) indicates that substantial inter-subject motor variability demands effective fusion of multi-distribution data for real-time, precise cross-subject and cross-task trajectory forecasting.
(c) reveals that human-exoskeleton systems are highly dynamic with coupled perception and control, calling for an efficient, robust, and real-time optimization scheme to ensure assistance performance. }. 
	\label{fig_pro}
\end{figure*}

\subsection{Problem Formulation}
Most state-of-the-art exoskeletons predominantly rely on single-step prediction \citep{molinaro2024estimating,molinaro2024task,luo2024experiment}, which involves learning a mapping from observations $\bb{o}$ to torque $\bb{a}$. 
Although the aforementioned methods have demonstrated promising assistance performance in complex tasks, they rely heavily on either expensive motion capture systems \citep{molinaro2024estimating,molinaro2024task} or sophisticated musculoskeletal models \citep{luo2024experiment,2025TAI}. Motion capture systems are impractical to deploy in dynamic outdoor environments (See Fig. \ref{fig_pro} (a)), while musculoskeletal models face substantial challenges in being trained for locomotion over complex terrains. Furthermore, single-step prediction suffers from inherent limitations, including network inference latency and high prediction variance, which inevitably degrade the comfort of robotic assistance.
Therefore, this paper aims to learn the two mapping functions defined in Eq. \eqref{map}, enabling user trajectory prediction in complex scenarios and subsequent torque output optimization, thereby delivering effective assistance.

While trajectory prediction for exoskeletons has been extensively investigated, its application to dynamic trajectory prediction and adaptive assistance in complex outdoor environments remains largely underexplored. 
Most existing works that apply trajectory prediction to practical exoskeleton assistance typically employ dynamic movement primitives (DMPs) \citep{chen2024learning}, probabilistic movement primitives (ProMPs) \citep{chen2024upper,dong2025}, or predefined functions \citep{lee2024ai} to reduce trajectory dimensionality and predict task-specific parameters. While these approaches have achieved efficient assistance performance in certain scenarios, they lack sufficient adaptability to handle highly dynamic and complex environments. Therefore, predicting high-dimensional trajectories in complex environments and integrating them into exoskeleton control remains an open and challenging problem.

\subsubsection{Learning the map function \texorpdfstring{$\mathcal{F}(\cdot)$}{F()}:}
Previous deep learning methods for exoskeletons, such as long short-term memory (LSTM), are often inadequate for high-dimensional long-term trajectory prediction. Furthermore, there exists significant inter-user variability in movement patterns and substantial distributional differences in multi-modal data \citep{liu2025weight} (See Fig. \ref{fig_pro} (b)). Directly training neural networks with deterministic policies under such conditions often leads to poor generalization performance. Conversely, learning a probabilistic policy $\mathcal{F}(\cdot)\sim p(\cdot|\mathbf{o}_t)\in[0,1]$ can enhance policy exploration and thus effectively mitigate this limitation. Generative approaches \citep{lipman2022flow,songdenoising}, which learn the probabilistic path from an initial distribution to a target distribution, have been widely adopted to achieve the above goal.

The human-exoskeleton system is a highly coupled and highly dynamic system, where the accuracy and real-time performance of trajectory prediction are of paramount importance. Generative methods, such as diffusion model \citep{songdenoising} and flow matching \citep{lipman2022flow}, start from a Gaussian distribution and iteratively generate the final target distribution through multiple steps. However, these methods incur substantial inference latency, rendering them unable to meet the stringent real-time prediction requirements of exoskeleton systems. For example, the diffusion model requires approximately $100$ denoising steps to generate a trajectory, and the inference time requires approximately $1$ s \citep{ding2025}. Furthermore, compared to FM, diffusion models exhibit significantly higher training complexity. Therefore, the first key challenge is how to effectively incorporate multi-modal data from diverse subjects into FM training, such that accurate trajectory prediction can be achieved while preserving stringent real-time inference performance (Solved in Section \ref{fm}).

\subsubsection{Learning the map function \texorpdfstring{$\mathcal{G}(\cdot)$}{G()}:}
Exoskeleton robots are designed to provide comfortable assistance during complex environmental tasks. Position control schemes are inherently insufficient to achieve this objective. In human-exoskeleton coupled systems, tracking dynamic trajectories using torque control schemes poses significant challenges (See Fig. \ref{fig_pro} (c)). 
This arises from the high dynamicity of human locomotion in complex scenarios, where predicted trajectories exhibit substantial deviations from actual human movements and optimized torque values oscillate significantly near zero. This, in turn, induces exoskeleton chattering and prevents the delivery of effective assistance. 
Accordingly, the second critical challenge is how to design an optimization objective function that can fully leverage trajectory information and ensure optimization stability, thereby optimizing torque outputs to achieve safe and robust assistance (Solved in Section \ref{control}).
	\section{Trajectory Prediction in Complex Environments via Fast Flow Matching}\label{fm}
This section begins with an introduction to the fundamental mathematical principles of flow matching. It then presents in detail the architecture and training procedure of the proposed fast FM algorithm, and concludes with a discussion of the deployment specifics, including the network framework and key design choices.

\subsection{Preliminaries for Flow Matching}
Given a sample $\mathbf{x}\in \mathbb{R}^d$ drawn from the known  distribution $p_{\text{src}}$ and a target sample $\mathbf{y}\in \mathbb{R}^d$ drawn from target distribution $p_{\text{tar}}$. Flow Matching (FM), a framework of a generative model, constructs a probability path $(p_t)_{0\leq t \leq 1}$ between the source distribution $p_{\text{src}}=p_0$ and the target distribution $p_{\text{tar}}=p_1$ \citep{albergo2022building}. FM can also build a time dependent flow $\psi_t(\mathbf{x}):[0,1]\times\mathbb{R}^d\xrightarrow{}\mathbb{R}^d$ between source data $\mathbf{x}$ and target data $\mathbf{y}$, where $\psi_t(\mathbf{x})\sim p_t$ and $\psi_0(\mathbf{x})=\mathbf{x}$. This flow can be determined by the following ordinary differential equation (ODE) \citep{lipman2024flow}:
\begin{equation}\label{eq_1}
    \frac{d\psi_t(\mathbf{x})}{dt}=v_t\left(\psi_t(\mathbf{x})\right), \quad t\in[0,1]
\end{equation}
where $v_t(\cdot):[0,1]\times\mathbb{R}^d\xrightarrow{}\mathbb{R}^d$ is the time dependent vector field, which is the core of FM and can control the direction of flow path, that is the vector field $v_t(\cdot)$ generates flow path $\psi_t(\cdot)$ (or probability path $p_t(\cdot)$) \citep{tong2023conditional}. The goal of FM is that the output of $\psi_1(\mathbf{x})$ is close to the expected value $\mathbf{y}$. Thus, $\mathbf{y} = \psi_1(\mathbf{x})$ can be regarded as the condition of flow (probability) path. A Gaussian conditional path $p_t(\cdot|\mathbf{y})=\mathcal{N}(\mu_t(\mathbf{y}),\sigma_t^2(\mathbf{y}))$ is usually employed to represent the conditional flow $\psi_t(\cdot|\mathbf{y})$ and inversely solve for the conditional vector field $v_t(\cdot|\mathbf{y})$ \citep{lipman2022flow}:
\begin{equation}
    \begin{aligned}
\psi_t(\mathbf{z}|\mathbf{y})&=\mu_t(\mathbf{y})+\mathbf{z}\cdot\sigma_t(\mathbf{y}),\\
v_t(\mathbf{z}|\mathbf{y})&=\frac{\sigma_t'(\mathbf{y})}{\sigma_t(\mathbf{y})}\left(\mathbf{z}-\mu_t(\mathbf{y})\right)+\mu_t'(\mathbf{y}),
    \end{aligned}
\end{equation}
where $\mu_t'(\cdot)$ and $\sigma_t'(\cdot)$ are the time derivation of mean $\mu_t(\cdot)$ and standard deviation $\sigma_t(\cdot)$, respectively. $\mathbf{z}\in \mathbb{R}^d$ is the auxiliary variable. Under the following conditions: $\mu_0(\mathbf{y})=\mathbf{x}$ and $\mu_1(\mathbf{y})=\mathbf{y}$, the mean $\mu_t(\cdot)$ and standard deviation $\sigma_t(\cdot)$ can be simplified to the following form \citep{tong2023conditional}:
\begin{equation}
    \mu_t(\mathbf{y})=(1-t)\mathbf{x}+t\mathbf{y},\quad \sigma_t(\mathbf{y})=\sigma,
\end{equation}
where the standard deviation is a small constant. Then, the conditional vector field $v_t(\cdot|\mathbf{y})$ can be derived as:
\begin{equation}\label{eq_4}
    v_t(\cdot|\mathbf{y})=\mathbf{y}-\mathbf{x}.
\end{equation}

During the prediction phase, the true target value $\mathbf{y}$ is not known a priori. We want to learn a time-dependent vector field $v_\theta(t, \mathbf{x})$ parameterized with $\theta$ based on the available training data $\mathcal{D}$. The parameters $\theta$ can be optimized by minimizing a flow matching loss:
\begin{equation}\label{eq_5}
    \mathcal{L}_{\text{FM}}(\theta)=\mathbb{E}_{t,p_t(\mathbf{z}|\mathbf{y})} \big\|v_\theta( \mathbf{x},t)-v_t(\mathbf{z}|\mathbf{y})\big\|^2_2,
\end{equation}
where $t\sim\mathcal{U}(0,1)$ and $\|\cdot\|^2_2$ is the squared 2-norm. This loss function is designed to enforce $v_\theta(t, \mathbf{x})$ to converge toward the ground-truth vector field (Eq. \eqref{eq_4}). Once $v_\theta(\mathbf{x},t)$ is obtained, given the sample $\psi_0(\mathbf{x})=\mathbf{x}$, the final output $\mathbf{y}=\psi_1(\mathbf{x})$ can be derived through $T$ iterative steps based on Eq. \eqref{eq_1}. The iteration equation can be represented as : $\psi_{t}(\mathbf{x})=\psi_{t-1}(\mathbf{x})+\Delta t \cdot v_{t-1}(\psi_{t-1}(\mathbf{x}))$, where $\Delta t=1/T$.

While a larger $T$ (finer discretizations of $[0,1]$) improves ODE solution accuracy, it increases inference time for the network-approximated $v_t$, degrading real-time generation performance. Consequently, improving the inference efficiency of FM is critical to robotic real-time control. Moreover, the generation $\mathbf{y}$ is based on $\mathbf{x}\sim p_0$, and in most existing works, $\mathbf{x}$ follows the distribution $\mathcal{N}(0,I)$. 
In practical robotic control, the motion trajectory $\mathbf{y}$ is generated using FM conditioned on the encoder $\mathbf{c}\in \mathbb{R}^d$ of the current observation. Thus, the generation of $\mathbf{y}$ should be guided by $\mathbf{c}$ to ensure task-aware trajectory synthesis. Therefore, effectively integrating observation into the FM framework presents another critical problem in robotic control applications.

\subsection{Fast FM Policy}
To solve the above two critical problems in robotic control for FM, we make two major modifications in the FM formulation: 1) adding the one-step trajectory generation error to the loss function accelerates inference while preserving prediction accuracy; 2) making the generation $\mathbf{y}$ (flow processes) conditioned on the encoder $\mathbf{c}$ of the input observation $\mathbf{o}$ to guide the vector filed training, where the condition can be denoted as $\mathbf{c}_t=\mathbf{E}(\mathbf{o}_t)$.
In exoskeleton assistance, the observation $\mathbf{o}\in \mathbb{R}^{T_o\times H_o}$ is multi-modal motion data measured from the IMU and motor, and the generation value $\mathbf{y}\in \mathbb{R}^{T_p\times 2H_p}$ ($T_p\times 2H_p=d$) is the joint trajectory. The proposed method aims to learn the policy distribution $p(\mathbf{y}|\mathbf{o})\in[0,1]$ through FM, that is, learning the map function $\mathcal{F}(\cdot)$. The following part provides details of the policy learning.
 
\subsubsection{Accelerating Inference:} FM generates $\mathbf{y}$ through $T$-step iterations, but it requires a certain inference time during actual prediction. Currently, the majority of acceleration methods for inference focus on reducing the number of iterations $T$, where $T$ typically denotes the number of function evaluations (NFE). In this work, we also adopt this perspective. For solving the $\psi_1(\mathbf{x})=\mathbf{y}$, if the vector field $v_t$ is known, Eq. \eqref{eq_1} admits a closed-form analytical solution:
\begin{equation}\label{eq_6}
\psi_t(\mathbf{x})=\psi_0(\mathbf{x})+\int_0^tv_\tau\big(\psi_\tau(\mathbf{x})\big)d\tau.
\end{equation}
where the true vector field $v_t$ is unknown, and the term $v_\theta(\mathbf{x},t)$ is trained to approximate the true vector field. Due to the highly dynamic nature of exoskeleton robots in outdoor assistive scenarios, this paper aims to train a one-step inference model (\textit{i.e.}, NFE $= 1$) to meet the stringent requirement for real-time performance. According to Eq. \eqref{eq_6}, the final predicted trajectory $\mathbf{y}$ can be expressed as
\begin{equation}\label{eq_7}
    \mathbf{\hat{y}}=\mathbf{x}+v_\theta(\mathbf{x},1).
\end{equation}

However, directly training the vector field using Eq. \eqref{eq_5} leads to significant error in one-step trajectory generation. In image generation, flow matching emphasizes similarity between images across different categories. In contrast, trajectory generation prioritizes predictive accuracy \citep{fu2025moflow}. To ensure high precision in one-step inference, we integrate the one-step trajectory prediction error into the training of the vector field $v_\theta(\mathbf{x},t)$, thereby guiding its optimization process. The corresponding loss function is formulated as follows:
\begin{equation}\label{eq_8}
    \mathcal{L}_\theta= \mathcal{L}_{\text{FM}}(\theta)+ \alpha\cdot D(\mathbf{y},\mathbf{\hat{y}}).
\end{equation}
where $\mathcal{L}_{\text{FM}}(\theta)$ and $\mathbf{\hat{y}}$ are defined in Eq. \eqref{eq_5} and Eq. \eqref{eq_7}, respectively. $D(\mathbf{y},\mathbf{\hat{y}})=\mathbb{E}_{\mathbf{x},\mathbf{y}}\|\mathbf{y}-\mathbf{\hat{y}}\|^2_2$ is a regularization term that measures the error between the true trajectory and one-step generation trajectory. The coefficient $\alpha\geq0$ is a hyperparameter used to trade off the prediction accuracy and generalization. A larger coefficient strengthens the constraint on trajectory accuracy, but it can impede the FM process from accurately learning the true trajectory distribution, thereby compromising the network's generalization ability. Conversely, an excessively small coefficient facilitates learning of the distribution, but often at the expense of reduced precision in the generated trajectories. Therefore, selecting an appropriate coefficient helps the velocity field maintain its ability to learn the trajectory distribution while further enhancing the precision of trajectory generation.

Once $v_\theta(\cdot)$ is trained, the desired trajectory can be accurately generated with a one step. Please note that the ground-truth trajectory $\mathbf{y}$ aligns with the observed values $\mathbf{o}$, and the predicted trajectory $\mathbf{\hat{y}}$ should also correspond to these observations. Below, we present the methodology for guiding trajectory generation $\mathbf{y}_t$ based on observed values $\mathbf{o}_t$, that is, learning the policy distribution $p(\mathbf{y}_t|\mathbf{o}_t)$.

\begin{figure}[!t]
	\centering
	\renewcommand{\algorithmicrequire}{\textbf{Input:}}
	\renewcommand{\algorithmicensure}{\textbf{Output:}}
	\removelatexerror
\begin{algorithm}[H]
	\caption{Fast FM Policy Training \& Inference}
	\label{alg1}
	\begin{algorithmic}[1]
    
\STATE \#\# \textbf{\texttt{Training}}
    
\REQUIRE Initial parameters $\theta$, prior distribution $p_0=\mathcal{N}(0,I)$, and the collected dataset $\mathcal{D}=\{\mathbf{o}_i,\mathbf{y}_i\}_{i=1}^N$. 
\ENSURE  Learned mean vector field parameters $\theta$.

\FOR {$i=1,2,\cdots, N_{\text{iter}} $}
       \STATE Sample $\mathbf{x}\sim p_0$, and sample time step $t$ from distribution $U$.
       \STATE Encode observation $\mathbf{o}_t$, and obtain condition $\mathbf{c}_t$.
       \STATE Calculate the target vector field $v_{tar}(\cdot)$ and the error between the true and one-step generation trajectory.
       \STATE Update $v_{\theta}(\cdot)$ through minimizing Eq. \eqref{eq_8}.
\ENDFOR
\STATE  \#\# \textbf{\texttt{One-step Inference}}
\REQUIRE The learned vector field $v_{\theta}$, the observation $\mathbf{o}$, and prior distribution $p_0$.
\ENSURE  Predicted trajectory $\mathbf{\hat{y}}$.
\STATE  Sample $\mathbf{x}\sim p_0$ and encode observation $\mathbf{c}=\mathbf{E}(\mathbf{o})$.
\STATE Calculate the predicted trajectory via Eq. \eqref{eq_7}.
\end{algorithmic}
\end{algorithm}
\end{figure}

\subsubsection{Guidance with Observation:} 
The generation of trajectories is influenced by the vector field $v_{\theta}(\cdot)$. This part illustrates how to incorporate the observation $\mathbf{o}$ into the training process of the vector field. The vector field guided by the observation $\mathbf{o}$ can be denoted as $v(\mathbf{x},t|\mathbf{c})$, where $\mathbf{c}$ is the encoder of the observation $\mathbf{o}$. From the above formulation, time $t$ can also be treated as a conditioning variable of the vector field. Here, both the observed values $\mathbf{o}$ and time $t$ are regarded as encoded inputs, and flow matching is utilized to decode the combined conditioning information along with the initial noise $\mathbf{x}$, thereby generating the final trajectory $\mathbf{y}$. Therefore, the trajectory generated under the guidance of observation $\mathbf{o}$ can be expressed as the following two components:
\begin{equation}\label{eq_10}
\begin{aligned}
        H_\text{enc}&=\texttt{EN}\big\{\mathbf{E}_t(t),\mathbf{E}_o(\mathbf{o})\big\},\\
        H_\text{dec}&=\texttt{DE}\big\{\mathbf{E}_x(\mathbf{x}),H_\text{enc}\big\}.
\end{aligned}
\end{equation}
where $\texttt{EN}(\cdot)$ and $\texttt{DE}(\cdot)$ represent the encoding of the conditioning inputs and the decoding of the trajectory, respectively. The encoder and decoder are both based on the transformer architecture. The details can be found in Section \ref{key_dec}. The terms $\mathbf{E}_t(\cdot)$, $\mathbf{E}_o(\cdot)$, and $\mathbf{E}_x(\cdot)$ represent the related encoders of time $t$, observation $\mathbf{o}$, and noise $\mathbf{x}$, respectively.

Algorithm \ref{alg1} summarizes the training and inference process of the fast FM policy. In the training phase, first, a sample $\mathbf{x}$ is drawn from the prior distribution $p_0$, while time step $t$ is sampled from a logit-normal distribution. Then, the observation $\mathbf{o}$ is encoded to obtain condition $\mathbf{c}$, and the error between the true trajectory and $K$ generated trajectories is calculated, which are both used to guide the trajectory generation. Finally, the target vector field $v_{tar}$ is computed, and then the parameter $\theta$ is updated via Eq. \eqref{eq_8}. In inference phase, once the observation-conditional vector filed $v_{\theta}(\cdot)$ is obtained, according to Eq. \eqref{eq_7}, the trajectory $\hat{\mathbf{y}}$ under the observation $\mathbf{o}$ can be obtained through one-step iteration, that is $\hat{\mathbf{y}}=\mathbf{x}+v_{\theta}(\mathbf{x},1|\mathbf{c})$, where $\mathbf{c}=\mathbf{E}(\mathbf{o})$ and $\mathbf{x}\sim \mathcal{N}(0,I)$. The trajectory also can be generated by $T$-step iterations in this setting, where $v(\mathbf{x},1)$ can be calculated by $\Delta t\sum_{i=0}^{T-1}v_{\theta}(\mathbf{x},t_i)$, and $t_i$ denotes the $i$-th time point uniformly discretized between $0$ and $1$ over $T$ intervals.

\begin{figure*}
	\centering
	\includegraphics[width=1.0\textwidth]{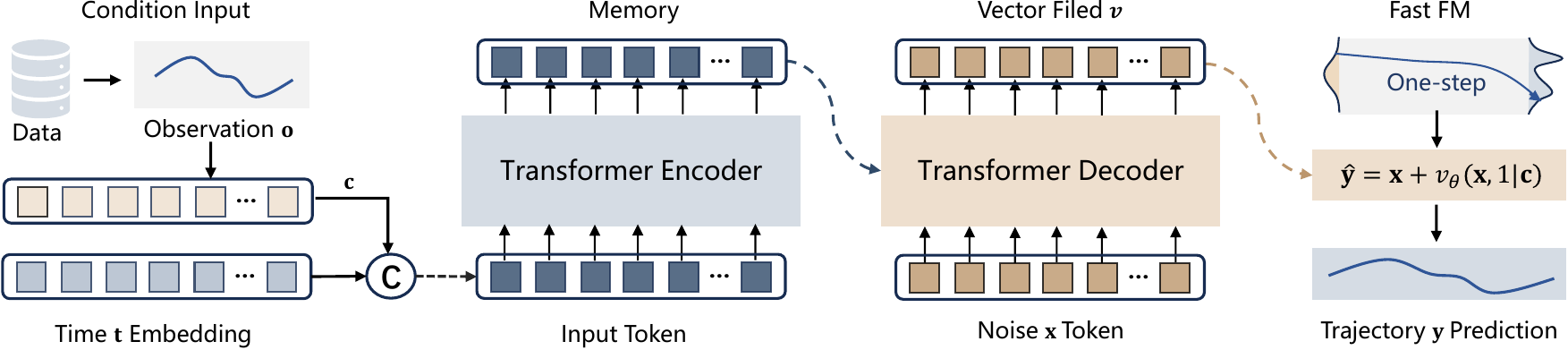}
    \captionsetup{justification=justified, singlelinecheck=false}
	\caption{Schematic of the fast FM network framework. The framework begins by encoding the observation $\mathbf{o}$ and time step $t$, which are then combined and fed into a transformer encoder module. The output $\mathbf{e}$ of the encoder, along with an encoding of the noise $\mathbf{x}$, serves as input to the transformer decoder module. The decoder outputs a vector field $v$, which is iterated in a one-step to generate the predicted trajectory $\mathbf{\hat{y}}$. }. 
	\label{fig2}
\end{figure*}

\subsection{Key Design Decisions}\label{key_dec}
In this part, we outline the key design decisions behind the fast FM policy and present its concrete implementation using transformer architectures. Fig. \ref{fig2} shows the overall schematic of the fast FM composition. Specifically, the detailed network architecture of the modules involved in Eq. \eqref{eq_10} is described as follows.
\subsubsection{Conditions Encoder:} The conditions for fast FM policy include two parts: the time $t$ and the observation $\mathbf{o}$. The observation $\mathbf{o}$ for trajectory generation consists of high-dimensional, multi-feature time-series data. To capture their temporal dependencies and fuse the multi-modal features, we employ a transformer-based architecture \citep{vaswani2017attention} for encoding the observational inputs. The core of the transformer can be represented as 
\begin{equation}
        \mathbf{E}(\mathbf{e}) = \texttt{softmax}\left(\frac{\mathbf{Q}(\mathbf{e})\mathbf{K}^T(\mathbf{e})}{\sqrt{d_k}}\right)\mathbf{V}(\mathbf{e}),
\end{equation}
where $\mathbf{Q}(\mathbf{e})$, $\mathbf{K}(\mathbf{e})$ and $\mathbf{V}(\mathbf{e})$ are query, key and value embedding results for input $\mathbf{e}$, respectively \citep{vaswani2017attention}, and $\sqrt{d_k}$ is scaling factor. For fast FM encode, the input $\mathbf{e}$ can be obtained through the following two parts: The time variable $t$ is encoded using sinusoidal positional encoding, and the observation values $\mathbf{o}$ are transformed into an embedded sequence via a shared multilayer perceptron. These two components are then combined to obtain $\mathbf{e}$, and then the term $\mathbf{e}$ is passed as input features to a transformer-based encoder for unified representation learning. 

The schematic diagram of the encoding process is shown in the first two subfigures of Fig. \ref{fig2}. The encoded values are subsequently passed to the trajectory decoding module to guide the direction of the flow matching-based trajectory generation. Note that in the conditional encoding process, the observation $\mathbf{o}$ is influenced by the observation horizon $T_o$, and the selection of the horizon parameter $T_o$ is critical. A minimal value of $T_o$ is detrimental to long-term prediction, while an overly large value increases inference time and may even degrade prediction performance. Therefore, selecting an appropriate horizon $T_o$ is essential to ensure inference speed while improving the accuracy of trajectory generation. For related ablation studies and analysis, please refer to Appendix \ref{other_results}. This stage of encoding observations $\mathbf{o}$ and temporal information $t$ lays the foundation for subsequent trajectory decoding.

\subsubsection{Trajectory Decoder:}
Trajectory decoding aims to generate the vector field $v$ based on the encoder context $\mathbf{c}$ and the input noisy $\mathbf{x}$, which is subsequently used to recover the trajectory $\mathbf{y}$ via a one-step iteration of the flow matching process.
Given that the trajectory itself is also high-dimensional data, we likewise adopt a transformer-based architecture to decode both the conditioned encoding $H_{\text{enc}}$ and the initial noise $\mathbf{x}$. The decoder incorporates a cross-attention mechanism to focus on the encoded conditioning input $H_{\text{enc}}$, and employs causal encoding to preserve temporal causality across time steps. During the trajectory generation process, the prediction horizon $T_p$ is a critical parameter. An excessively large value of $T_p$ can lead to degraded prediction accuracy, whereas an overly small value may fail to meet the assistance requirements of the exoskeleton. To address this, we generate a trajectory with a longer length $T_p$, but only the first $L$ time steps of the predicted trajectory are selected as the control target for the exoskeleton. This approach maintains prediction accuracy while simultaneously fulfilling the need for long-term assistance. Appendix \ref{other_results} gives the specific details regarding the selection of parameters $L$ and $T_p$.

\section{Torque Optimization via Model Predictive Control}\label{control}

The previous section utilizes fast FM to generate joint trajectories $\mathbf{y}_t=\{\bb{y}_{t+i},\bb{\dot{y}}_{t+i}\}_{i=1}^{T_p}$, which provide joint angle $\bb{y}$ and angular velocity $\bb{\dot{y}}$ information. This section aims to design an appropriate optimization objective to optimize the torque sequence $\mathbf{a}_t=\{\mathbf{u}_{t+i}\}_{i=0}^{N-1}$ based on the predicted trajectory, thereby providing comfortable and compliant assistance and achieving efficient human–robot assistance in complex environments. Model predictive control (MPC), which is based on the optimization of long-term future trajectories, offers an excellent optimization framework. On this basis, torque optimization is carried out in the following.

\subsection{Dynamic Model for Exoskeleton}
MPC relies on the dynamic model of the exoskeleton robot. Let the number of joints be $n$. Here, $n$ is equivalent to the aforementioned $H_p$ in representing the number of joints. For brevity, $n$ is used uniformly hereafter.
Following \citet{han2023human}, the dynamic model of an exoskeleton robot, with consideration of human–robot interaction $\mathbf{T}_{int}$, can be given by
\begin{equation}\label{eq_11}
        \mathbf{M}\ddot{\bb{y}}+\mathbf{C}(\bb{y},\dot{\bb{y}})\dot{\bb{y}}+\mathbf{G}(\bb{y})+\mathbf{T}_{int}=\mathbf{u},
\end{equation}
where $\bb{y}\in \mathbb{R}^{n}$ and $\bb{y}_d\in \mathbb{R}^{n}$ are the actual and desired joint angle, respectively. $\mathbf{M}\in \mathbb{R}^{n\times n}$ and $\mathbf{C}(\bb{y},\dot{\bb{y}})\in \mathbb{R}^{n\times n}$ are the $n \times n$ inertial and velocity-dependent matrices, $\mathbf{G}(\bb{y})\in \mathbb{R}^{n}$ is the $n$-dimensional gravitational torque vector, and $\mathbf{u} \in \mathbb{R}^{n}$ and $\mathbf{T}_{int}\in \mathbb{R}^{n}$ are motor output torque and the interaction torque, respectively. This model has the properties that $\mathbf{M}$ is symmetric and positive definite, and that $\mathbf{M}-2\mathbf{C}$ is skew-symmetric \citep{2017yu}. The parameters for $\mathbf{M}$, $\mathbf{C}$, and $\mathbf{G}$ are obtained from the Lagrangian dynamics and are functions of the exoskeleton link's mass, inertia, and center of mass, which can be acquired from the CAD model. 

In general, the interaction torque $\mathbf{T}_{int}$ is represented as impedance model \citep{chen2024upper}, that is $\mathbf{B}(\bb{y}_d-\bb{y})+\mathbf{K}(\dot{\bb{y}}_d-\dot{\bb{y}})=\mathbf{T}_{int}$, where $\mathbf{B}\in \mathbb{R}^{n\times n}$ and $\mathbf{K}\in \mathbb{R}^{n\times n}$ are stiffness and damping matrices of impedance model. Nevertheless, the impedance model shares a critical flaw with PID control: significant discrepancies between $\bb{y}_d$ (the predicted trajectory) and the real trajectory cause large fluctuations in the calculated interaction torque, thereby compromising the effectiveness of the torque optimization. In complex operational scenarios, the impedance model exhibits poor applicability in control frameworks built upon high-dynamic trajectory prediction.

Crucially, the optimal exoskeleton assistance does not correspond to the smallest possible interaction torque; instead, it requires the interaction torque to be proportional to the true human-generated torque \citep{molinaro2024task}. Since the impedance model cannot adequately represent human motor biomechanics, we integrate human kinematic information into the interaction torque estimation framework while ensuring the stability of the optimization process. Inspired by the work in \citep{8643426}, which employs joint angles $\bb{y}$ and angular velocities $\bb{\dot{y}}$ to fit joint torques, we adopt the same two kinematic variables to approximate the human–robot interaction torques, that is:
\begin{equation}\label{eq_13}
    \mathbf{T}_{int} = \mathbf{T}_0 +K_0\bb{y}_d +K_1\bb{\dot{y}}_d,
\end{equation}
where $\mathbf{T}_0$, $K_0$ and $K_1$ are the fitting coefficients. Here, we utilize the predicted trajectory $\mathbf{y}_d$ instead of the actual trajectory $\mathbf{y}$. This design choice is motivated by the need to mitigate the detrimental impact that large oscillations in the measured trajectory would otherwise exert on the downstream optimization procedure.

Let the state be $\mathbf{s}=[\bb{y};\dot{\bb{y}}]\in \mathbb{R}^{2n}$ and action be $\mathbf{u}\in \mathbb{R}^{n}$. Then, based on Eq. \eqref{eq_11}, the dynamic relationship between the derivation of the state $\mathbf{\dot{s}}$ and action $\mathbf{u}$ can be written as
\begin{equation}
    \mathbf{\dot{s}}=\begin{bmatrix} \bb{s}_1 \\ \mathbf{M}^{-1}\big(\mathbf{u-\mathbf{C}}\bb{s}_1+\mathbf{G}(\bb{s}_0)-\mathbf{T}_{int}(\bb{s}_0,\bb{s}_1)\big) \\ \end{bmatrix},
\end{equation}
where $\bb{s}_0=\bb{y}$ and $\bb{s}_1=\mathbf{\dot{y}}$, and the next state can be denoted as $\mathbf{s}_{t+1}=\mathbf{s}_{t}+\mathbf{\dot{s}}_t\Delta t$. Then, the discrete-time system with state $\mathbf{s}_{t}$ and control input $\mathbf{u}_t$ can be characterized by a deterministic transition function $f$ that maps the current state-action pair to the next state:
\begin{equation}
    \mathbf{s}_{t+1}=f(\mathbf{s}_{t},\mathbf{u}_t).
\end{equation}
The above equation can be used to constrain the MPC optimization problem. The details of torque optimization can be found in the part below.

\begin{figure*}
	\centering
	\includegraphics[width=1\textwidth]{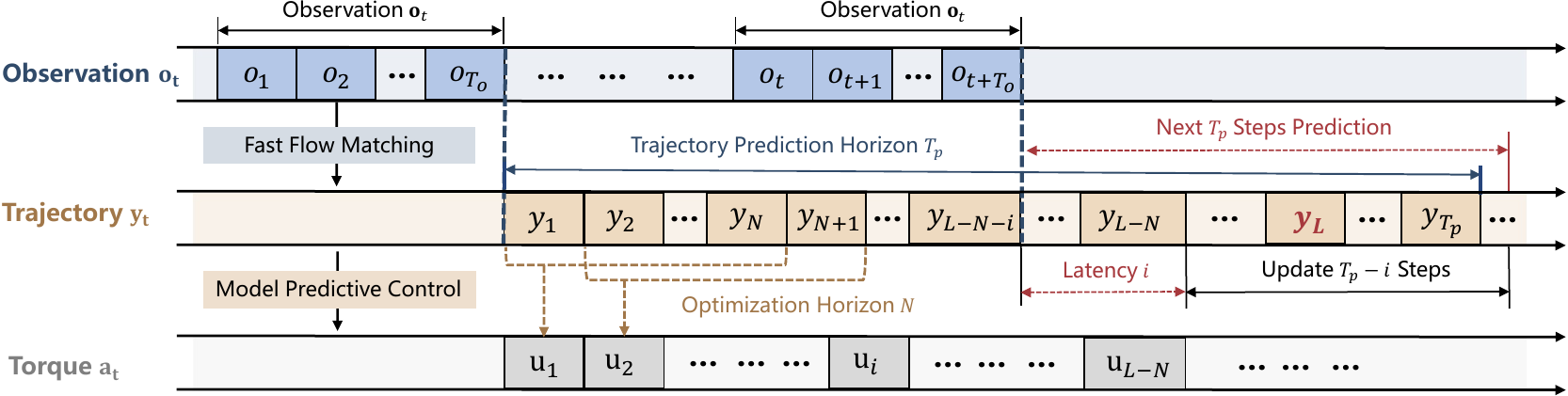}
    \captionsetup{justification=justified, singlelinecheck=false}
	\caption{Control data flow diagram of the proposed \texttt{ExoTraj} policy. First, the fast FM is used to predict the $T_p$-step-ahead trajectory using the observations $\mathbf{o}$ acquired over the past $T_o$ time steps, where the first $L$ steps of the trajectory are used to optimize torque. Then, MPC is adopted to optimize the current torque based on the trajectory spanning from the current time step to the $N$-step-ahead horizon. A new round of trajectory prediction is triggered when the receding horizon progresses to the ($L-N-i$)-th time step (where $i$ denotes the number of time steps consumed by algorithm inference), and the trajectory of the remaining $T_p-i$ steps after the ($L-N$)-th step is updated simultaneously.}. 
	\label{fig3}
\end{figure*}

\subsection{Objective Problem Formulation}

When tracking a desired joint trajectory $\mathbf{y}_d$ by regulating motor torque with conventional PID control, the resulting torque profile often exhibits undesirable fluctuations and poor tracking accuracy, which may ultimately degrade the overall assistance performance. In contrast, MPC incorporates long-term dynamic behavior into the optimization process and can optimize torque based on the designed objective:
\begin{equation}\label{eq_14}
\begin{cases}
        \min\limits_{\mathbf{a_t}}&\sum_{i=0}^{N-1}\bb{\ell}(\mathbf{s}_{t+i},\mathbf{u}_{t+i})+\bb{\ell}_N(\mathbf{s}_{t+N}),\\
        \text{s.t.} & \mathbf{s}_{t+1}=f(\mathbf{s}_{t},\mathbf{u}_t), \quad \mathbf{s}_{t}=\mathbf{s}_{init}\\
        &h(\mathbf{s_t},\mathbf{u_t})\leq \mathbf{0}, \quad \forall t\in \mathbf{Z}^{+}\\
\end{cases}
\end{equation}
where $\ell(\cdot)$ and $\ell_N(\cdot)$ are running and final cost function, respectively, $h(\mathbf{s_t},\mathbf{u_t})$ are the constrain function of the state and action, and $\mathbf{s}_{init}$ is the initial state. The term $N$ refers to the horizon of optimization, which is a key hyperparameter for MPC performance. The final cost function $\ell_N(\cdot)$ can be represented as $\Delta\mathbf{s}_{t+N}^\mathrm{T}\mathbf{O}\Delta\mathbf{s}_{t+N}$, where $\mathbf{O}\succeq 0\in \mathbb{R}^{2n\times 2n}$ is final weight matrices. For exoskeleton assistance, the running cost function $\ell(\cdot)$ consists of three components: trajectory tracking error ($\Delta_1$), torque safety ($\Delta_2$), and torque smoothness ($\Delta_3$). The first two terms aim to minimize the trajectory tracking error and constrain the joint torque within a safe range, which can be mathematically expressed as:
\begin{equation}\label{eq_17}
\Delta_1+\Delta_2=\Delta\mathbf{s}_t^\mathrm{T}\mathbf{Q}\Delta\mathbf{s}_t+\mathbf{u}_t^\mathrm{T}\mathbf{R}\mathbf{u}_t
\end{equation}
where $\Delta\mathbf{s}_t=\mathbf{s}_t-\mathbf{s}_{t,\text{ref}}$, and $\mathbf{Q}\succeq 0\in \mathbb{R}^{2n\times 2n}$ and $\mathbf{R}\succeq 0\in \mathbb{R}^{n\times n}$ are tracking error and torque weighting matrices, respectively. The third term $\Delta_3$ enforces the smoothness of the torque, which requires that the torque gradually changes between consecutive time steps to ensure a comfortable experience and natural assistance. This is achieved by penalizing the rate of change of the joint torque, mathematically expressed as:
\begin{equation}\label{eq_18}
    \Delta_3 = \left(\frac{\mathrm{d}\mathbf{u}_t}{\mathrm{d}t}\right)^{\mathrm{T}} \mathbf{P}\left(\frac{\mathrm{d}\mathbf{u}_t}{\mathrm{d}t}\right),
\end{equation}
where $\mathbf{P}\succeq 0\in \mathbb{R}^{n\times n}$ is the weighting matrices. Considering discrete torque form $\mathbf{a}_t=[\mathbf{u}_{t};...;\mathbf{u}_{t+N-1}]\in\mathbb{R}^{N n\times1}$, the time derivative of $\mathbf{a}_t$ is expressed as \citep{2025Diff}
\begin{equation}\label{eq_19}
\begin{aligned}
        \frac{\mathrm{d}\mathbf{a}_t}{\mathrm{d}t}&=\left[\frac{\mathrm{d}\mathbf{u}_{t}}{\mathrm{d}t};...;\frac{\mathrm{d}\mathbf{u}_{t+N-2}}{\mathrm{d}t}\right]\in \mathbb{R}^{n(N-1)\times 1}\\
        &=\frac{1}{\Delta t}\left[\mathbf{u}_{t+1}-\mathbf{u}_{t};...;\mathbf{u}_{t+N-1}-\mathbf{u}_{t+N-2}\right]\\
        &=\frac{1}{\Delta t}\begin{bmatrix}
-1 & 1 & & & \\
& -1 & 1 & & \\
& & \ddots & \ddots & \\
& & & -1 & 1
\end{bmatrix}\mathbf{a}_t\\
& = \mathbf{A}\mathbf{a}_t, \text{ where } \mathbf{A}\in\mathbb{R}^{n(N-1)\times Nn}
\end{aligned}
\end{equation}

Then, substituting Eqs. \eqref{eq_17}-\eqref{eq_19} into Eq. \eqref{eq_14}, the objective can be written as
\begin{equation}\label{eq_20}
\begin{cases}
\min\limits_{\mathbf{a}_t} &\mathbf{e}_t^\mathrm{T}\mathbf{U}\mathbf{e}_t+\mathbf{a}_t^\mathrm{T}\mathbf{V}\mathbf{a}_t,\\
\text{s.t.} & \mathbf{s}_{t+1}=f(\mathbf{s}_{t},\mathbf{u}_t),\\
& \mathbf{s}_{t}\in[\mathbf{s}^l,\mathbf{s}^u],\mathbf{u}_{t}\in[\mathbf{u}^l,\mathbf{u}^u],
\end{cases}
\end{equation}
where $\mathbf{e}_t=[\Delta\mathbf{s}_t;...;\Delta\mathbf{s}_{t+N}]\in\mathbb{R}^{2n(N+1)\times 1 }$ and $\Delta\mathbf{s}_t=\mathbf{s}_t-\Delta\mathbf{s}_{t,\text{ref}}$. $\mathbf{s}^l$ and $\mathbf{s}^u$ and $\mathbf{a}^l$ and $\mathbf{a}^u$ are the low and high bounds of state and action, respectively. The matrices $\mathbf{U}\succeq 0\in \mathbb{R}^{2n(N+1)\times 2n(N+1)}$ and $\mathbf{V}\succeq 0\in \mathbb{R}^{Nn\times Nn}$ can be represented as:
\begin{equation*}
    \begin{aligned}
        &\mathbf{U}=\diag\{\mathbf{Q}_1,...,\mathbf{Q}_N,\mathbf{O}\},\\
        &\mathbf{V}=\diag\{\mathbf{R}_1,...,\mathbf{R}_N\}+\begin{bmatrix}\mathbf{A}\\ \mathbf{0}\end{bmatrix}^\mathrm{T}\diag\{\mathbf{P}_1,...,\mathbf{P}_N\}\begin{bmatrix}\mathbf{A}\\ \mathbf{0}\end{bmatrix},
    \end{aligned}
\end{equation*}
where $\mathbf{Q}_i=\mathbf{Q}$, $\mathbf{R}_i=\mathbf{R}$, $\mathbf{P}_i=\mathbf{P}$ and $\mathbf{0}\in\mathbb{R}^{n\times Nn}$. Thus, the torque can be optimized using Eq. \eqref{eq_20}. Note that each optimization uses trajectories from the current to $N$ future steps for tracking, but only the first optimized torque value is applied to the motors, which are re-optimized at every time step. By designing the optimization objective under the predictive control paradigm of the model, we achieve the construction of the mapping $\mathcal{G}(\cdot)$ from the reference trajectories to the corresponding output control torques.


\subsection{Control Framework of \texttt{ExoTraj}}
By combining the trajectory generated by fast FM and the MPC-based optimization, this paper proposes a general exoskeleton assistance policy,  denoted as \texttt{ExoTraj}, for complex environments. \texttt{ExoTraj} employs a hierarchical control architecture, where the high controller utilizes FM to rapidly generate task-specific trajectories $\mathbf{y}$, while the low controller leverages MPC to compute motor torque command $\mathbf{a}$. 
Fig. \ref{fig3} illustrates the pipeline of \texttt{ExoTraj} for converting observations $\mathbf{o}$ into control signals $\mathbf{a}$ within the control process. Firstly, based on the past $T_o$ time steps of observations, FM is employed to generate trajectories for the next $T_p$ time steps. The first $L$ predicted steps are utilized for torque optimization, and MPC with a horizon of $N$ is applied to compute the optimal torque output. This cyclic workflow enables real-time and high-efficiency assistance for the exoskeleton system.
To ensure the MPC horizon remains $N$ steps even toward the end of the $L$-step execution, the prediction horizon $T_p$ is set longer than $L$. 

Real-time performance is critical for exoskeleton robot assistance. For the high-level controller, inference is performed on an edge computing GPU device, with an inference time of approximately $5-6$ ms. For the low-level controller, torque optimization is implemented using the CasADi toolkit, requiring around $2$ ms. The system operates at a sampling and control command output frequency of $200$ Hz. However, the inference time occasionally exceeds $5$ ms. To ensure real-time model inference, this paper introduces an inference interval $i$, which is when the system reaches step $L-N-i$, a new trajectory is generated for the next $T_p$ steps. 
Please note that although network inference is triggered at time step $L-N-i$, the subsequent trajectory is only used at time step $L-N$. 
Moreover, for the MPC torque optimization from steps $L-i$ to $L$, the input trajectory consisting of $N$ points is still derived from the previously generated trajectory of $T_p$ points. The analysis of the effect of the usage trajectory steps $L$ on continuous trajectory prediction accuracy is presented in Appendix \ref{ablation_L}. 


\begin{figure*}
	\centering
	\includegraphics[width=1.0\textwidth]{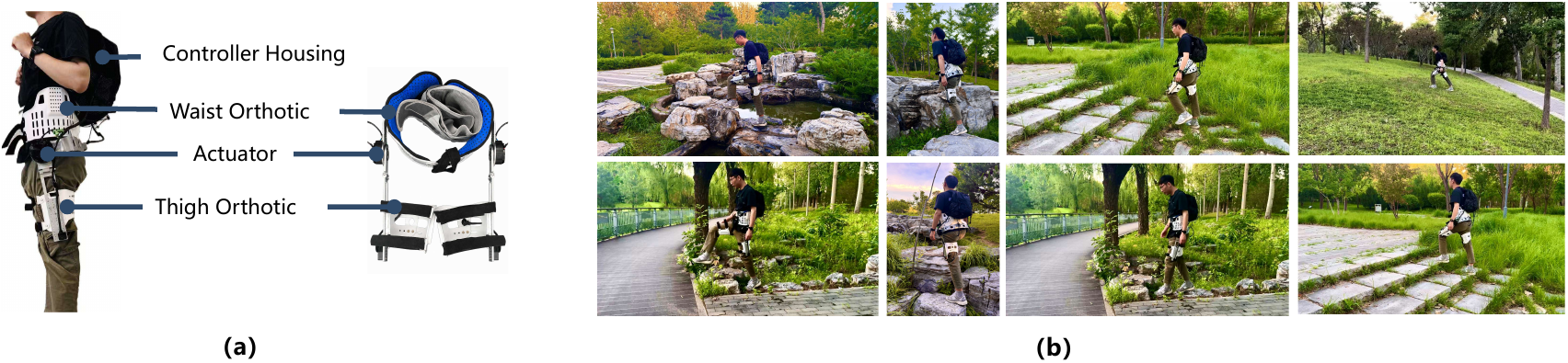}
    \captionsetup{justification=justified, singlelinecheck=false}
	\caption{(a) illustrates the structure and system configuration of the proposed untethered hip exoskeleton, consisting of a waist-mounted battery pack, drive motors, and orthotic supports for the waist and thighs. (b) presents a schematic diagram of the terrain in complex outdoor environments for exoskeleton assistance, including rugged terrain, rocky slopes, and uneven lawns.}. 
	\label{fig4}
\end{figure*}

\section{Experiment Setup}\label{setup}
This section details the experimental setup, which comprises the architecture of the exoskeleton system, the protocol for data collection and experiment, and the configuration of critical control parameters.

\subsection{Details of Experiment Procedures}
Fig. \ref{fig4} (a) illustrates the mechanical configuration and core components of the hip exoskeleton. The system consists of a waist harness, adjustable elastic belts, thigh support frames, controller housing, and hip joint actuation modules, with a total mass of $6.8$ kg. This device provides assistance for hip flexion and extension motions, offering a continuous torque of $9$ Nm and a peak torque of $22$ Nm.
Torque assistance is achieved through a quasi-direct drive actuator (AK80-9, T-Motor, China). Three inertial measurement units (MPU9250) are mounted on bilateral thighs and the lower torso to capture human locomotion kinematics. The edge computing device (NVIDIA Jetson Orin NX) undertakes real-time neural network inference and torque optimization. The microcontroller (STM32F429) is responsible for sensor data sampling and motor drive control.
The exoskeleton is firmly fastened to the user’s pelvis and thighs via customized orthoses. A two-degree-of-freedom passive revolute joint preserves natural hip mobility during movement. More details can be found in Appendix \ref{app_composition}.

Based on the established hip exoskeleton platform, the experiments are divided into offline and online stages. The offline stage focuses on multimodal data acquisition and assistance policy training, whereas the online stage verifies the actual assistive performance under complex, unstructured outdoor scenarios.
In the offline phase, $12$ healthy adult participants are recruited for data collection. The first $8$ participants (AB01-AB08) are instructed to wear the exoskeleton system configured in a non-assistive passive mode, and to walk naturally at a range of self-selected walking speeds in a dynamic outdoor walking environment (depicted in Fig. \ref{fig4} (b)). The predefined outdoor walking course covers typical complex unstructured terrains, including sloped slopes of variable gradient, rough uneven ground, and other challenging outdoor surfaces. During the walking trials, full multi-modal sensor data from the exoskeleton system are synchronously recorded. The remaining four subjects are employed for cross-subject evaluation of exoskeleton assistance performance. More details can be found in Appendix \ref{dataset}.

In the online phase, to verify the real-time assistance performance of \texttt{ExoTraj}, participants wore the active exoskeleton system and completed walking trials in the complex outdoor environment. The cross-subject experimental design is adopted with three experimental conditions: 1) No exo: natural overground walking without exoskeleton wear; 2) Assist off: walking with the exoskeleton configured in zero-torque passive mode; 3) Assist on: walking with the exoskeleton enabled with the proposed assistance policy.
To quantitatively evaluate the assistance efficiency of the proposed policy, three categories of physiological metrics are synchronously measured and recorded for all trials: 1) Metabolic rate, calculated by indirect calorimetry based on oxygen consumption and carbon dioxide production measurements obtained with the COSMED K5 device; 2) Muscle activation, acquired via a surface electromyography (EMG) system (ELONXI EMG-C4); 3) Heart rate, measured using a high-precision Garmin wearable heart rate monitor.
The experiments adopt a cross-subject evaluation paradigm, in which test subjects' motion data remain unseen during policy training.


\subsection{Parameters Configuration}
This part presents the specific parameter configurations used in the high-level (trajectory prediction) network and the lower-level MPC controller. For exoskeleton control, the parameters of the dynamic model, $\mathbf{M}(\bb{y})$, $\mathbf{C}({\bb{y}},\dot{\bb{y}})$, and $\mathbf{G}(\bb{y})$ in Eq. \eqref{eq_11}, and the MPC parameters optimization horizon $N$, error weight matrix $\mathbf{Q}$, torque weight matrix $\mathbf{R}$, and penalization weight matrix $\mathbf{P}$ should be tuned according to the specific exoskeleton model and control objectives. The detailed parameter configuration is provided in Table \ref{Configuration}.
In the MPC formulation, the state $\mathbf{s}$ (angle $\bb{y}$ and angular velocity $\bb{\dot{y}}$) and action $\mathbf{a}$ constraints are defined with lower and upper bounds of $\mathbf{s}^l$ and $\mathbf{s}^u$, set to values $[-0.4\pi,-100,-15]$ and $[0.4\pi,100,15]$, respectively. The error matrix $\mathbf{Q}$, torque matrix $\mathbf{R}$, penalization weight matrix $\mathbf{P}$, and terminal weight matrix $\mathbf{O}$ are set as $\text{diag}\{10\bb{I}_n,0.1\bb{I}_n\}$, $0.1\bb{I}_n$, $0.001\bb{I}_n$ and $\bb{I}_{2n}$, respectively, where $\bb{I}_i$ denotes the $i$-dimensional identity matrix. 
For the interaction torque estimation $\mathbf{T}_{int}$ described in Eq. \eqref{eq_13}, the three coefficients $\mathbf{T}_0$, $K_0$ and $K_1$ are configured as $0.01\bb{I}_n$, $\bb{0}$, and $4\bb{I}_n$, respectively. 
The prediction horizon $N$ of the MPC torque optimization is set to $15$, with an average per-iteration computational time of about $2$ ms, which fully meets the system’s real-time control requirements.

\begin{table}
	\normalsize
	\caption{Parameters configuration for the policy \texttt{ExoTraj}.}
	\label{Configuration}
	\centering
	\resizebox{8.5cm}{!}{
        \begin{tabular}{cc|cc}
\toprule[1pt] 
\multicolumn{2}{c|}{\textbf{High-Level Controller}}&\multicolumn{2}{c}{\textbf{Lower-Level Controller }}\\
Parameter & Value & Parameter & Value \\
\midrule
Batch size & $512$      & Inertial mat.$\mathbf{M}$ & $0.0156\bb{I}$\\
Training epoch & $200$    &Velocity mat.$\mathbf{C}$ & $\bb{0}$\\
Learning rate & $10^{-4}$     & Gravity mat.$\mathbf{G}$ & $0.879\sin (\bb{y})$\\
Coefficient $\alpha$ & $1$ &Error mat.$\mathbf{Q}$ & $\diag\{10,0.1\}\bb{I}$    \\
Obs. horizon $T_o$ & $250$ & Torque mat.$\mathbf{R}$ & $0.5\bb{I}$\\
Pred. horizon $T_p$ & $40$ & Penal. mat. $\mathbf{P}$ & $0.001\bb{I}$\\
Infer. latency $i$ & $1$ & Terminal mat. $\mathbf{O}$ & $0.1\bb{I}$\\
Usage steps $L$ & $20$ & Opt. horizon $N$ & $15$\\
\toprule[1pt] 
\end{tabular}
}
\end{table}

The network parameters of fast FM consist of a transformer-based network architecture and associated training hyperparameters (See Table \ref{Configuration}). The transformer network is configured with $4$ attention layers, each employing $4$ attention heads. The linear layer contains $256$ neurons. The model is trained with a learning rate of $1\times10^{-4}$, a batch size of $256$, for $200$ epochs. The time $t$ is sampled from the normal logit distribution. The weight coefficient $\alpha$ is set to $1$. The input observation features $\mathbf{o}$ have a dimension of $[250, 20]$, and the output trajectory $\mathbf{y}$ is represented as a $[40,4]$ feature vector, that is, the observation horizon $T_o=250$ and the prediction horizon $T_p=40$. The entire network comprises approximately $6$ million parameters.
In the deployed system, both the predicted joint trajectories and the sensor observations (IMU and joint angles) supplied to the network are updated at $5$ ms intervals. The high-level controller requires approximately $5-6$ ms for trajectory prediction, while the low-level MPC optimization takes $2$ ms. 
To ensure real-time performance, we introduce an inference latency compensation step $i$. A $1$-step-ahead prediction strategy is adopted, which starts prediction $5$ ms in advance to mitigate the inference latency. In this scenario, a new cycle of trajectory prediction is triggered when the receding horizon progresses to the ($L-N-i$)-th time step. The details of data flow can be found in Fig. \ref{fig3} and Fig. \ref{figa1}.

	\section{Experimental Results}\label{Experiments}
\begin{table*}
	\normalsize
	\caption{Comparison results of prediction performance for various methods in the offline and online phase. Offline evaluation uses the in-distribution (\texttt{InDist}) dataset and the cross-subject (\texttt{CrossSub}) dataset, while online tests adopt exoskeleton-based real-time prediction. Bold and underlined values denote the best and second-best performance, respectively.}
	\label{performance}
	\centering
	\resizebox{17.5cm}{!}{
        \begin{tabular}{c|cc|cc|ccc|ccc}
\toprule[1pt] 
\multirow{3}{*}{\textbf{Methods}} & \multicolumn{7}{c|}{\textbf{Offline Phase}}&\multicolumn{3}{c}{\textbf{Online Phase}}\\
& \multicolumn{2}{c|}{{\texttt{InDist} ($T_p=20$)}} & \multicolumn{2}{c|}{{\texttt{InDist} ($T_p=40$)}}& \multicolumn{3}{c|}{{\texttt{CrossSub} ($T_p=20$)}} & \multicolumn{3}{c}{{Usage steps $L=20$}}\\
& ADE &$R^2(\%)$&ADE &$R^2(\%)$& ADE &$R^2(\%)$& Time (ms)& ADE &$R^2$(\%)& Time (ms)\\
\midrule
LSTM & $\underline{0.008}$ & $\mathbf{99.95}$ 
& $\underline{0.011}$ & $\underline{99.93}$
& $0.026$ & $95.42$& $4.24$
& $0.758$ & $70.13$ & $23.23$\\
TCN & $0.014$ & $99.29$ 
& $0.024$ & $98.90$
& $0.024$ & $96.65$& $\mathbf{1.91}$
& $0.546$ & $87.37$ & $\mathbf{2.53}$\\
TF & $0.025$ & $99.50$
& $0.030$ & $99.36$
& $0.030$ & $94.75$ & $\underline{1.95}$
& $\underline{0.515}$ & $\underline{88.35}$ & $\underline{3.48}$\\
FM & $0.010$ & $99.88$ 
& $0.012$ & $99.87$
& $\underline{0.023}$ & $\underline{96.70}$& $12.70$
& $0.611$ & $82.00$ & $13.85$\\
\midrule
Fast FM & $\mathbf{0.004}$ & $\underline{99.94}$ 
& $\mathbf{0.006}$ & $\mathbf{99.93}$
& $\mathbf{0.018}$ & $\mathbf{98.10}$& $4.10$
& $\mathbf{0.443}$ & $\mathbf{89.30}$ & $5.09$\\
Improved & $\downarrow44.3\% $ & $\downarrow0.0\%$ 
& $\downarrow44.4\%$ & $\uparrow0.0\%$
& $\downarrow20.5\%$ & $\uparrow1.5\%$& $\uparrow2.2$
& $\downarrow14.0\%$ & $\uparrow1.1\%$ & $\uparrow2.6$\\
\toprule[1pt] 
\end{tabular}
}
\end{table*}

This section validates the effectiveness of the proposed fast FM-based unified control policy in highly dynamic environments through extensive offline and online experiments, aiming to address the following research questions:
\begin{itemize}[leftmargin=*] 
    \item $\mathbf{Q}_1:$ For the high-level controller, how does the proposed trajectory prediction method compare with other data-driven methods?
    \item $\mathbf{Q}_2:$ For the low-level control module, what are the advantages of using model predictive control over traditional control methods?
    \item $\mathbf{Q}_3:$ What is the effectiveness of the assistance of the proposed \texttt{ExoTraj} policy in complex and dynamic environments?
\end{itemize}

The following experimental section will address each of these questions in turn. Detailed parameter analyses, ablation studies, and additional experimental results are presented in Appendix \ref{other_results}.
Related project materials, including movies and documentation, are available at \href{https://xiaoyinliu0714.github.io/Home_ExoTraj/}{Page}.

\subsection{Performance of Trajectory Prediction (\texorpdfstring{$Q_1$}{Q1})}
\subsubsection{Evaluation Metrics:} Fast FM predicts the future motion trajectory $\mathbf{y}$ over a horizon of $T_p$ times, based on the observed exoskeleton state $\mathbf{o}$ from the past $T_o$ times. The performance of trajectory prediction can be quantified using two metrics: the average displacement error (ADE) and the squared R ($R^2$). A smaller ADE value, approaching zero, indicates lower trajectory prediction error, while an $R^2$ value closer to $1$ signifies better prediction performance from a statistical perspective. The equations can be represented as:
\begin{equation}\label{pred_acc}
\begin{aligned}
     ADE=\frac{1}{N_p}\frac{1}{T_p}\sum_{i=1}^{N_p}\sum_{t=1}^{T_p}\left\|\bb{{y}}_{i,t}-\bb{\hat{y}}_{i,t}\right\|_2,\\
     R^2 = 1-\frac{\sum_{i=1}^{N_p}\sum_{t=1}^{T_p}\left\|\bb{{y}}_{i,t}-\bb{\hat{y}}_{i,t}\right\|_2}{\sum_{i=1}^{N_p}\sum_{t=1}^{T_p}\left\|\bb{{y}}_{i,t}-\bb{\bar{y}}_{i,t}\right\|_2},
\end{aligned}
\end{equation}
where $N_p$ is the number of trajectory, $\bb{{y}}_{i,t}$ represents the the motion angles of the two joints at the $t$-th time step of the $i$-th trajectory, $\|\cdot\|_2$ denotes the $L_2$-norm, and $\bb{\bar{y}}_{i,t}$ is the ground truth position of the center point, which can be calculated by $\sum_i\sum_{t}\bb{y}_{i,t}/N_pT_p$. We evaluate the algorithm's performance in both offline and online phases.
In the offline phase, we employ the cross-validation protocol on the collected dataset: data from $8$ subjects are used for model training, while data from the remaining $4$ subject are held out exclusively to evaluate the model's generalization capability.
In the online phase, system performance is quantified by computing the trajectory error between the network's real-time predictions (generated from streaming sensor data) and the corresponding ground-truth trajectories, recorded while human subjects operate the exoskeleton.

\subsubsection{Offline Evaluation:} To demonstrate the predictive accuracy and generalization of fast FM in the offline phase, we compare it with representative data-driven approaches: long short-term memory (LSTM) \citep{hochreiter1997long}, temporal convolutional network (TCN) \citep{lea2017temporal}, transformer (TF) \citep{vaswani2017attention}, and flow matching (FM) \citep{lipman2022flow}. We re-implemented the code and conducted evaluations on the two different types of datasets: the first is an in-distribution (\texttt{InDist}) test set, where $5$\% of the data from the first $8$ subjects are held out and remain completely unseen during model training; the second is a cross-subject (\texttt{CrossSub}) test set, consisting of locomotion data collected from entirely novel subjects who did not participate in the training phase. Table \ref{performance} compares the performance of various methods in the above two offline datasets, in terms of prediction metrics defined in Eq. \eqref{pred_acc} for $20$ and $40$ prediction steps ($T_p$) given $250$ past observations, and corresponding network inference times (unit: ms). The last row indicates the performance improvement rate of the best method over the second-best one or the degradation rate relative to the optimal value.

Table \ref{performance} demonstrates that fast FM achieves the highest prediction accuracy on in-distribution (\texttt{InDist}) validation data when prediction horizons $T_p=20$ and $T_p=40$, with about $44.0\%$ reduction in prediction error (ADE). While conventional neural networks exhibit fast inference speeds, their performance in long-term prediction tasks is comparatively inferior. The standard FM algorithm delivers superior prediction accuracy after three iterations, yet it suffers from significantly slower inference latency. 
On the cross-subject (\texttt{CrossSub}) dataset (for generalization performance evaluation), all methods exhibit a substantial degradation in prediction performance. Among them, LSTM experiences the most pronounced drop in generalization capability, whereas FM maintains better predictive performance. Fast FM still retains a favorable generalization prediction performance. When prediction horizon \(T_p\) equals $20$, its prediction error declines by $20.5\%$ (ADE).
Overall, compared with other approaches, fast FM achieves competitive prediction accuracy and generalization capability while guaranteeing real-time inference.

\begin{figure*}
	\centering
    \begin{minipage}{1\textwidth}
    \centering
    \includegraphics[width=\linewidth]{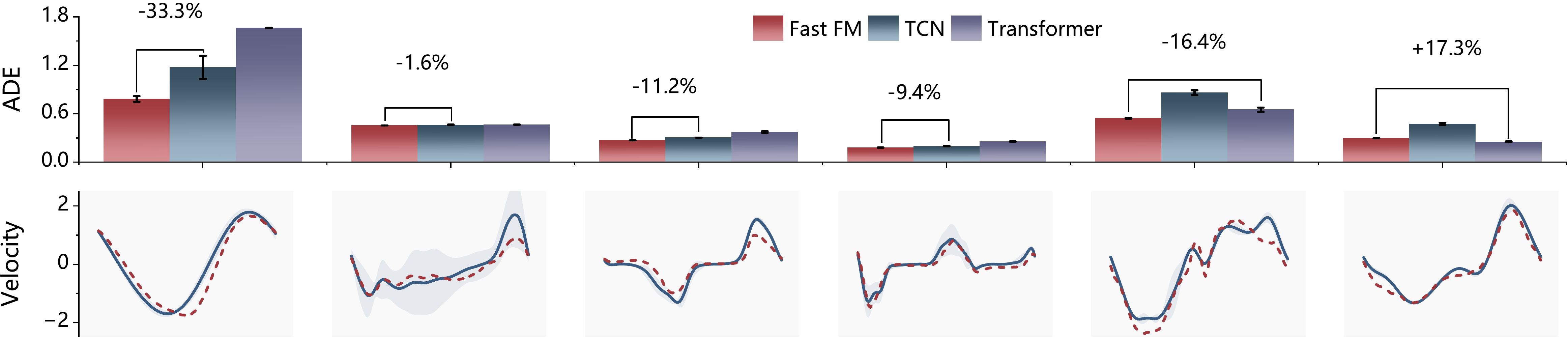}
  \end{minipage}
  \hfill
  \begin{minipage}{1\textwidth}
    \centering
    \includegraphics[width=\linewidth]{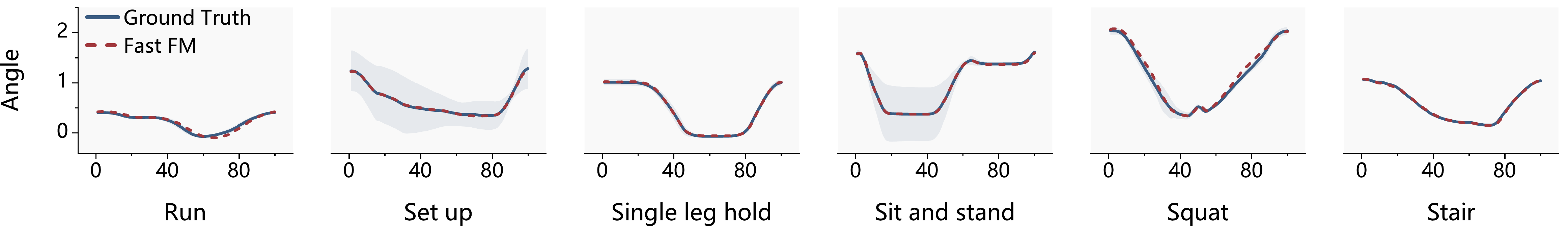}
  \end{minipage}
    \captionsetup{justification=justified, singlelinecheck=false}
	\caption{ Comparison of online trajectory prediction among Fast FM, TCN, and Transformer for six activities: running, obstacle crossing, leg swing pause, sit-to-stand, squatting, and stair climbing. Each full motion corresponds to one gait cycle, and all curves are scaled to $0$-$100$ cycles, and the true angular velocity is filtered with a cutoff frequency of $5$ Hz.}. 
	\label{fig6}
\end{figure*}

\subsubsection{Online Evaluation:} To verify the performance of fast FM in the real-time online phase, the test is conducted on an unseen subject, whose motion data are completely excluded from the network training pipeline. The subjects are instructed to perform a series of complex motor tasks, including several previously unseen actions not present in the training dataset, such as rapid deep squats, single-leg suspension, abrupt stopping, backward locomotion, and high-obstacle crossing. The trained model is converted to the TensorRT format and deployed on edge computing devices for real-time inference. During walking trials, both the predicted and true trajectories are recorded synchronously. In the online prediction phase, the prediction horizon is set to \(T_p=40\) and the actual usage trajectory length is \(L=20\). 
Table \ref{performance} presents the online real-time trajectory prediction accuracy of different methods across subjects. This shows that fast FM still achieves the optimal prediction performance, with the maximum prediction error reduced by $14.0\%$. Both LSTM and FM methods exhibit significant prediction errors, primarily due to their excessive online inference latency, which causes a substantial temporal misalignment between the predicted trajectories and the ground-truth trajectories.
It can be observed that the prediction errors of all methods generally increase in the online phase compared to the offline phase, which is primarily attributed to the relatively large prediction errors in angular velocity. Table \ref{angle_velocity} presents the prediction accuracies of angle and angular velocity separately during the online prediction stage. This shows that all methods achieve relatively high prediction accuracy for the angle. For angular velocity, due to its larger value range and higher dynamic characteristics than angle, an average displacement error (ADE) of around $0.7$ is within the acceptable range.

\begin{table}
	\normalsize
	\caption{Comparison results of prediction accuracy for angular trajectory and angular velocity trajectory in the online phase.}
	\label{angle_velocity}
	\centering
	\resizebox{8.5cm}{!}{
        \begin{tabular}{c|cc|cc|cc}
\toprule[1pt] 
\multirow{2}{*}{\textbf{Methods}} & \multicolumn{2}{c|}{\textbf{Angle}}&\multicolumn{2}{c|}{\textbf{Angle Velocity}}&\multicolumn{2}{c}{\textbf{Combined}}\\
& ADE &$R^2$&ADE &$R^2$& ADE &$R^2$\\
\midrule
LSTM & ${0.037}$ & ${99.3}$ 
& ${0.756}$ & ${66.1}$
& $0.758$ & $70.1$\\
TCN & $\mathbf{0.022}$ & $\underline{99.5}$ 
& $0.545$ & $86.9$
& $0.546$ & $87.4$\\
TF & $0.052$ & $98.7$
& $\underline{0.510}$ & $\underline{88.0}$
& $\underline{0.515}$ & $\underline{88.3}$ \\
FM & $0.037$ & $99.3$ 
& $0.608$ & $79.9$
& ${0.611}$ & ${82.0}$\\
\midrule
Fast FM & $\underline{0.024}$ & $\mathbf{99.7}$ 
& $\mathbf{0.442}$ & $\mathbf{88.1}$
& $\mathbf{0.443}$ & $\mathbf{89.3}$\\
\toprule[1pt] 
\end{tabular}
}
\end{table}

Fig. \ref{fig6} compares the real-time trajectory prediction performance of different methods for human subjects wearing the exoskeleton across various tasks. The comparison is conducted across three methods: TCN, Transformer, and the proposed fast FM. Note that the exoskeleton is set to zero-torque mode during this test phase; because LSTM and FM incur high inference latency and introduce prediction errors, these two methods are not included in the comparison. 
In the above experiments, each subject performs a single type of movement repeatedly. One full execution of a movement is defined as one cycle, and all joint angle trajectories are scaled to the range of $0$–$100$.
The bar chart in Fig. \ref{fig6} shows that fast FM achieves superior alignment and higher prediction accuracy across different movements. For highly dynamic movements, the maximum prediction error (ADE) is reduced by up to $33.3\%$ and $53.0\%$ (run movement) compared with TCN and Transformer, respectively. The average error of fast FM is reduced by $21.7\%$ and $18.6\%$ for TCN and Transformer, respectively. Fig. \ref{fig6} also presents the ground-truth trajectories and predicted trajectories for the specified motor tasks. For angle trajectories, the predicted values closely match the ground truth. For angular velocity, the predicted values show deviations in the peak regions, but the overall trends are basically consistent.
The above experimental results show that fast FM still achieves consistently favorable prediction performance in complex dynamic scenarios, outperforming the other two baseline methods, confirming its feasibility for practical deployment.

\subsection{Performance of Torque Optimization (\texorpdfstring{$Q_2$}{Q2})}
\subsubsection{Evaluation Metrics:} The performance of torque optimization is evaluated via two core metrics: assistance comfort and robustness. 
Robustness is reflected in the amplitude of fluctuations in the optimized torque output under significant perturbations or adversarial attacks on the predicted trajectory.
Assistance comfort can be indirectly characterized by torque smoothness and the consistency between torque and joint angle trajectories, where consistency refers to the degree of proximity between the gait cycle phase of the trajectory peak and that of the zero-torque point. 
This peak point corresponds to the stance-to-swing phase transition point in the gait cycle. Since the swing phase is characterized by high angular velocities, even a minor lead or lag in the assistive torque will result in significant user discomfort. Therefore, achieving perfect temporal consistency between the gait trajectory and the generated torque at this specific point is of paramount importance.
In this paper, the Jerk window can quantify the smoothness of the torque curve. Given the $N_T$ discrete torque points $\{\mathbf{u}_i\}_{i=0}^{N_T}$, the formula to calculate Jerk is presented as follows.
\begin{equation}\label{eq_22}
    S_\text{rms} = \sqrt{\frac{1}{N_T-3}\sum_{i=1}^{N_T-3}\|S_i\|^2},
\end{equation}
where $S_i=\mathbf{u}_{i+3}-3\mathbf{u}_{i+2}+3\mathbf{u}_{i+1}-3\mathbf{u}_{i}$. 
Given the joint trajectory \(\mathbf{y}\) and torque profile \(\mathbf{a}\) within a single gait cycle, we can derive the gait phase position of the trajectory peak \(\Phi_y\) and the gait phase position of the zero-torque point \(\Phi_a\), based on which the consistency metric can be defined as:
\begin{equation}\label{T_shif}
    T_\text{shif} = |\Phi_y-\Phi_a|,
\end{equation}
Then, combining Eq. \eqref{eq_22} and Eq. \eqref{T_shif}, the comfort metric of assistance can be indirectly measured by
\begin{equation}\label{comfort}
    C_\text{mpc} = \lambda_r\cdot\exp\{-\omega_r S_\text{rms}\}+\lambda_t\cdot\exp\{-\omega_t T_\text{shift}\},
\end{equation}
where $\omega_r=0.01$ and $\omega_t=1$ are scale factors, $\lambda_r=0.5$ and $\lambda_t=0.5$ are weight coefficients, and \(C_{\text{mpc}} \in [0,1]\), with values closer to $1$ indicating better assistance comfort and values closer to $0$ indicating worse comfort.
The robustness of assistance $R_\text{mpc}$ is quantified by measuring the variations in assistance torque profiles and their smoothness characteristics under baseline conditions and after introducing additive noise or targeted adversarial attacks. Therefore, we define the metric below to measure the robustness of the control method:
\begin{equation}
\begin{aligned}
    R_\text{mpc} = &\lambda_a\cdot\exp\{-\omega_a\cdot\left\|\mathbf{a}^d-\mathbf{a}^p\right\|_2\}+\\
    &\lambda_c\cdot\exp\{-\omega_c\cdot\left\|C^d_\text{mpc}-C^p_\text{mpc}\right\|_2\},
    \end{aligned}
\end{equation}   
where $\omega_a=0.01$ and $\omega_c=1$ are scale factors, $\lambda_a=0.6$ and $\lambda_c=0.4$ are weight coefficients, $\mathbf{a}^p$ and $\mathbf{a}^d$ correspond to the torque optimization profiles obtained before and after disturbance injection, respectively, and $C^p_\text{rms}$ and $C^d_\text{rms}$ denote the smoothness performance indices before and after disturbance noise injection, respectively. A value of $R_\text{mpc}$ closer to $1$ implies stronger robustness, while a smaller value indicates poorer robustness.

\begin{figure*}
	\centering
	\includegraphics[width=1\textwidth]{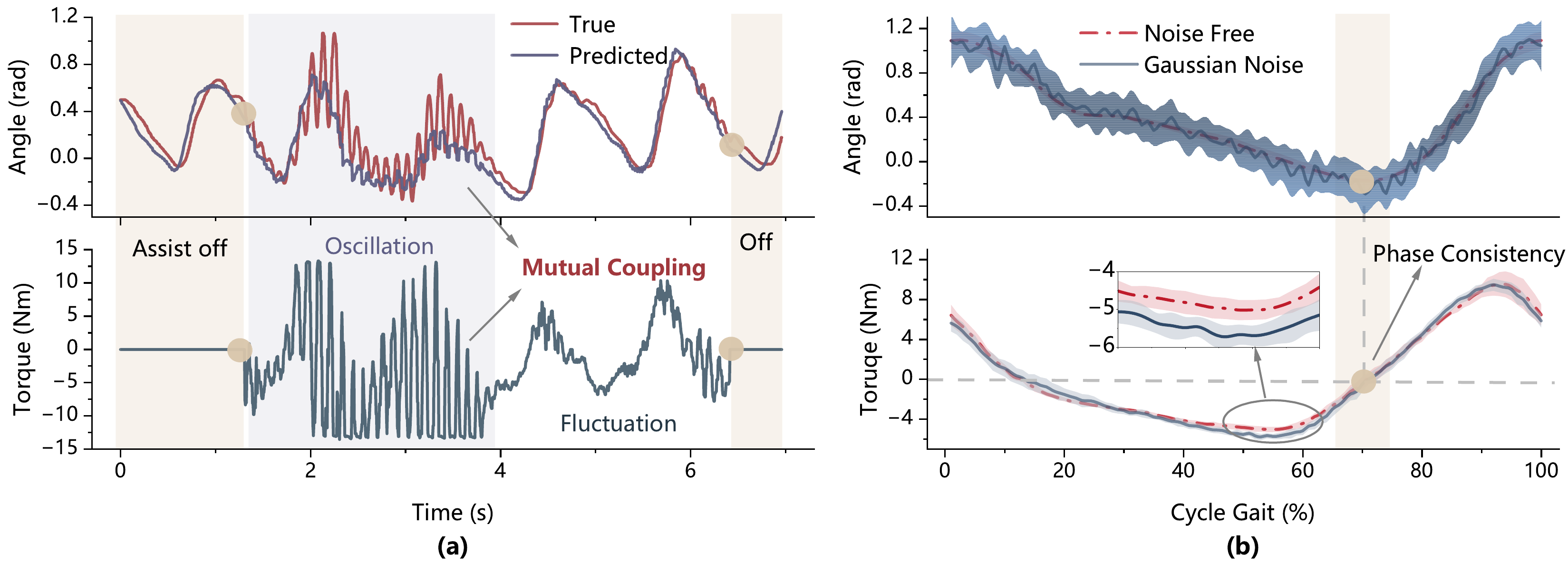}
    \captionsetup{justification=justified, singlelinecheck=false}
	\caption{Torque optimization comparison results. (a) illustrates the torque optimization performance of the PID controller. (b) compares the optimized torque profiles obtained via MPC optimization before and after injecting Gaussian noise into the predicted trajectories. In each subplot, the upper panel depicts the reference trajectory, while the lower panel presents the corresponding optimized torque. Shaded regions indicate the variance across multiple trials.}. 
	\label{fig7}
\end{figure*}

\subsubsection{Comfort:} Based on the predicted trajectories, we compare the assistance comfort metrics between the conventional PID algorithm and our MPC algorithm during the assistance process. Fig. \ref{fig7} (a) gives the optimized torque profiles generated by the conventional PID control method during treadmill walking at a constant speed of $1.0$ m/s. 
As PID control operates on an error-driven principle, we predict the reference trajectory $20$ steps ahead, corresponding to a $100$ ms time horizon. The control torque is derived from the error between the predicted state at the $20$-th future step and the current measured state. A PD control law is adopted herein, with a proportional gain of $20$ and a derivative gain of $0.005$.
Assist off indicates that the exoskeleton is not providing any assistance. It is verified that the $20$-step-ahead reference trajectory can be accurately predicted under the unassisted condition. However, once the exoskeleton engages in active assistance, the system exhibits undesirable oscillations.
This phenomenon stems from the inherent coupling between the exoskeleton's perception and control loops. The PID controller's performance is highly sensitive to trajectory prediction accuracy. Any significant instantaneous prediction error generates antagonistic control torques, which exacerbate prediction errors and ultimately cause erroneous control outputs in subsequent cycles. Even when the reference trajectory prediction is sufficiently accurate and the system remains free of oscillations, the assistive torque profile still exhibits undesirable fluctuations, which significantly compromise the human-exoskeleton interaction comfort.

In contrast, the MPC-optimized assistance torque exhibits significantly lower sensitivity to trajectory prediction errors and demonstrates superior robustness. Fig. \ref{fig7} (b) presents the predicted reference trajectory (Noise Free) and assistance torque optimized by MPC over a single gait cycle. This shows that the MPC-optimized torque exhibits excellent smoothness and closely follows the intrinsic kinematic pattern of human locomotion. Specifically, the generated torque decays to zero and undergoes full polarity reversal precisely at the trajectory peak amplitude, consistent with the natural kinematic characteristics of human limb motion.
Based on Eqs. \eqref{eq_22}–\eqref{comfort}, the MPC-optimized controller achieves torque smoothness and phase consistency metrics of $8.38$ and $0.02$, respectively, corresponding to a comprehensive assistance comfort metric of $0.95$. In contrast, the PID-optimized controller yields metrics of $15.76$ (torque smoothness) and $0.09$ (phase consistency), with a lower comprehensive comfort score of $0.88$. This further validates the inherent advantages of the MPC-based optimization scheme.


\subsubsection{Robustness:}
In practical human-in-the-loop assistance, predicted trajectories suffer from fluctuations and intrinsic human gait variability, requiring MPC optimization to be robust to disturbances and uncertainties. This part compares the performance of the optimized torque before and after noise injection/adversarial attacks. All experiments are conducted on a constant-speed treadmill to ensure consistent baseline conditions and predicted trajectories across trials. Fig. \ref{fig7} (b) presents the results of the optimized torque profiles after random noise injection into the predicted trajectories, where the applied noise follows a zero-mean Gaussian distribution $\mathcal{N}(0.2)$ with variance $0.2$. This figure shows that noise injection introduces significant fluctuations to the model-predicted trajectories. Yet, the optimized torque remains nearly unchanged, which indirectly corroborates the robustness of the MPC based on the optimization objective in Eq. \eqref{eq_20}.
Table \ref{robustness} presents the quantitative results of the robustness performance metrics under two noise injection modalities (Gaussian noise $\mathcal{N}(\cdot)$ and impulse noise $\mathcal{P}(\cdot)$) with different intensities. Gaussian noise is added to every time step of predicted trajectories, and impulse noise is randomly inserted at a single time step. To ensure a fair comparison, the subject is instructed to walk on a treadmill at a constant speed of $1.0$ m/s. Performance metrics are presented as mean and variance across multiple gait cycles.
Table \ref{robustness} shows the robustness index \(R_{\text{mpc}}\) presents a moderate decreasing trend with the rise of noise intensity, while still staying close to the noise-free reference level. Upon the introduction of noise disturbances, the \(C_{\text{mpc}}\) metric exhibits minimal variation, confirming that the optimized torque maintains comparable smoothness and consistency with and without noise. The maximum \(L_2\) norm of the difference between the optimized torque and the nominal torque remains approximately $10$; these discrepancies are primarily attributed to inter-subject variations in gait cycle duration at the specified walking speed, which introduce spurious torque offsets during the temporal scaling process, as illustrated in Fig. \ref{fig7} (b). Overall, the above results sufficiently verify that the developed MPC-based optimization framework possesses excellent intrinsic robustness.

\begin{table}
	\normalsize
	\caption{Comparative results of the robustness performance metrics under different types of noise injection for cross-subject. Results are presented as the mean and standard deviation across multiple gait cycle segments.}
	\label{robustness} 
	\centering
	\resizebox{8.4cm}{!}{
		\begin{tabular}{cc|ccc}  
			\toprule[1pt] 
			\multicolumn{2}{c|}{\multirow{2}{*}{\textbf{Attack Type}}}&\multicolumn{3}{c}{\textbf{Robustness Metrics}} \\
             && $\|\mathbf{a}^d-\mathbf{a}^p\|_2$ &$C_\text{mpc}$ & $R_\text{mpc}$ \\ 
			\midrule  
			\multirow{3}{*}{\rotatebox{90}{\textbf{Gauss.}}}& $\mathcal{N}(0.1)$
            &$\mathbf{8.24\pm3.74}$ &$0.95\pm0.01$ &$\mathbf{0.95\pm0.02}$\\
			&$\mathcal{N}(0.2)$ &$8.99\pm2.22$ &$0.95\pm0.00$ &$0.95\pm0.01$\\
			&$\mathcal{N}(0.5)$ & $12.27\pm2.98$& $0.94\pm0.01$&$0.93\pm0.02$\\ 
            \midrule
            \multirow{3}{*}{\rotatebox{90}{\textbf{Pulse}}} &$\mathcal{P}(0.5)$&$\mathbf{6.86\pm1.11}$ & $0.95\pm0.00$&$\mathbf{0.96\pm0.01}$\\
			&$\mathcal{P}(1)$ &$9.63\pm2.74$ & $0.94\pm0.00$&$0.94\pm0.01$\\
			&$\mathcal{P}(2)$ &$9.24\pm1.70$ & $0.95\pm0.00$&$0.94\pm0.01$\\ 
            \midrule
            \multicolumn{2}{c|}{\textbf{Noise Free}} &$0.00\pm0.00$ & $0.95\pm0.00$&$\mathbf{1.00\pm0.00}$\\  
			\bottomrule[1pt]  
		\end{tabular}
	}
\end{table}

\begin{figure*}
	\centering
	\includegraphics[width=1\textwidth]{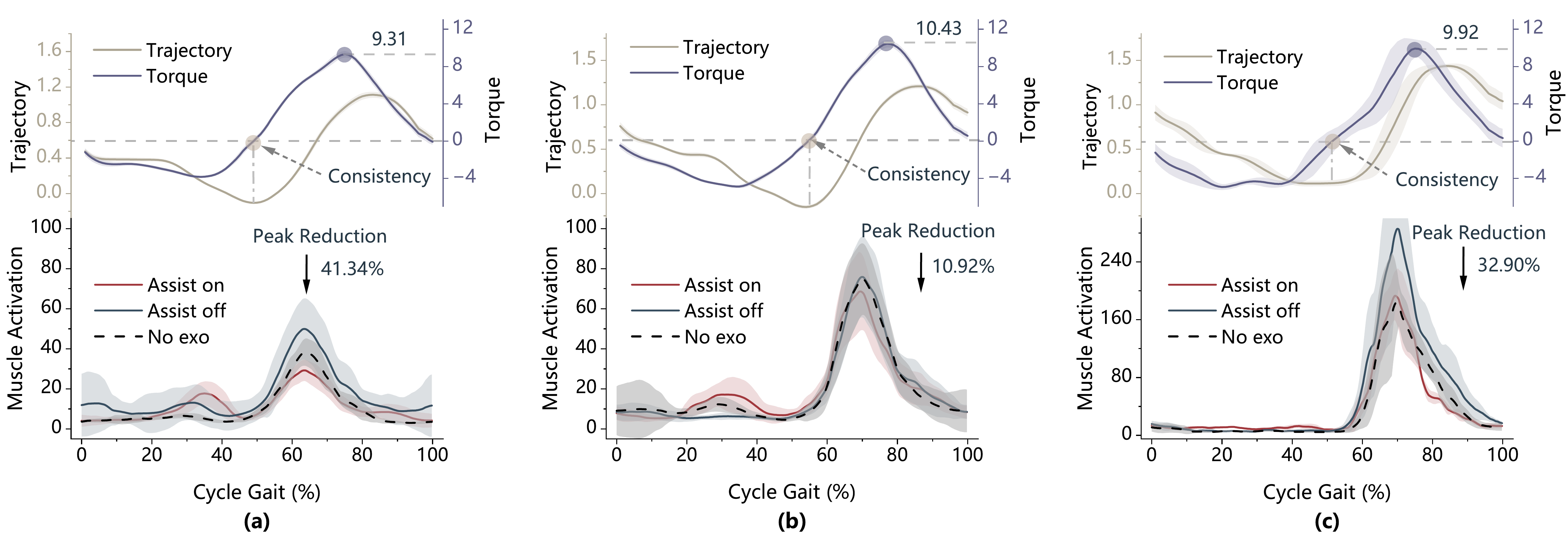}
    \captionsetup{justification=justified, singlelinecheck=false}
	\caption{ Comparison of muscle activation profiles under three assistance conditions (Assist on, Assist off, and No exo). Subplots (a), (b), and (c) present the results obtained on level ground, ramp, and stair locomotion, respectively. In each subplot, the upper trace shows the true trajectory and optimized torque, and the lower trace presents the corresponding muscle activation results.}. 
	\label{fig9}
\end{figure*}

\begin{figure*}
	\centering
	\includegraphics[width=1\textwidth]{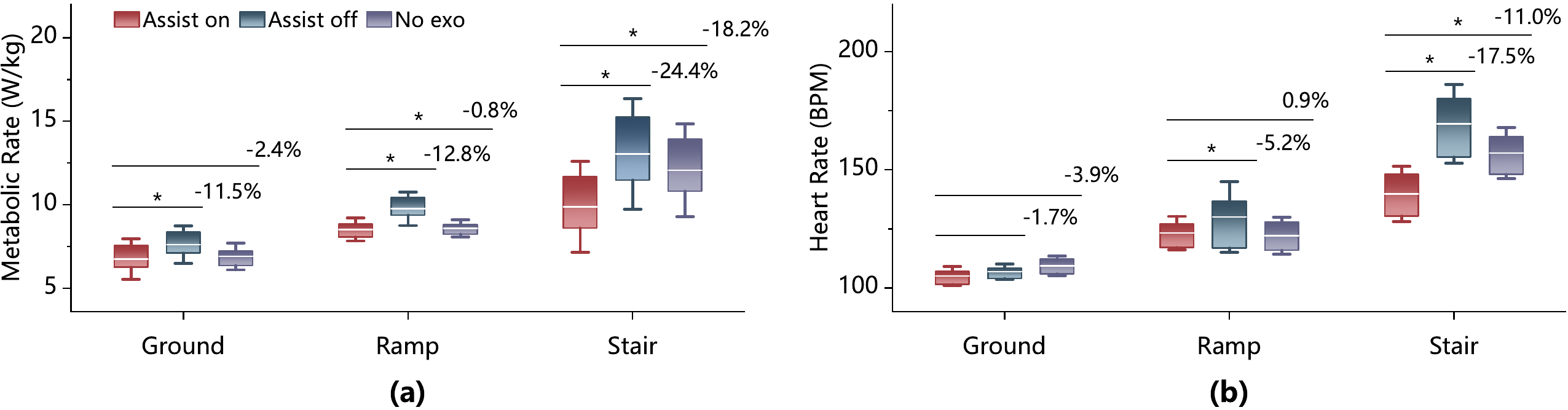}
    \captionsetup{justification=justified, singlelinecheck=false}
	\caption{Performance comparison results of metabolic rate (a) and heart rate (b) under three assistance conditions across three terrain types: level ground, sloped terrain, and staircase.}. 
	\label{fig10}
\end{figure*}

\subsection{Assistance Performance (\texorpdfstring{$Q_3$}{Q3})}
\subsubsection{Evaluation Metrics:} The efficacy of exoskeleton assistance can be quantified via changes in human physiological metrics, such as muscle activation level $A_t$ (unit: \si{\micro\volt}), metabolic rate $P_t$ (unit: W/kg), and heart rate $H_t$ (unit: BPM). We compare the effects of three conditions, No Exoskeleton (No exo), Exoskeleton without Assistance (Assist off), and Exoskeleton with Assistance (Assist on), on the three metrics across different scenarios.
In muscle activation experiments, the root mean square (RMS) value of the EMG signal \(E_t\) is used to quantify the level of muscle activation. The corresponding formula is given by: \citep{winters2012multiple}:
\begin{equation}
    A_t=\sqrt{1/T\sum\nolimits_{t=1}^TE_t^2},
\end{equation}
where $T$ denotes the duration of the time window.
For the metabolic rate trials, after collecting oxygen consumption (${V}O_2$) and carbon dioxide production (${V}CO_2$) data, the instantaneous metabolic power $P_t$ normalized by body mass $m$ is calculated as \citep{molinaro2024estimating}:
\begin{equation}
P_t=\frac{0.278\cdot {V}O_2 +0.075 \cdot {V}CO_2}{m}.
\end{equation}
All subsequent experiments adopt an experimental paradigm for cross-subjects.

\subsubsection{Muscle Activation Change:} Due to the high sensitivity of EMG sensors to motion interference, cross-subject muscle activation experiments are conducted under three relatively stationary conditions: level ground, a slope, and a staircase. Specifically, a walking speed of $1.25$ m/s is maintained on both level and inclined surfaces (\ang{10}), achieved using a motorized treadmill. During stair climbing, subjects walk at a relatively constant speed of approximately $50$ steps per minute. The final muscle activation profile is obtained by averaging the activation across multiple gait cycles. 
Fig. \ref{fig9} presents the muscle activation and the corresponding torque averaged over multiple gait cycles across three terrain conditions under three distinct assistance modes: Assist on, Assist off, and No exo. 
These results demonstrate that the \texttt{ExoTraj} policy yields substantial reductions in muscle activation:
1) Relative to the Assist off condition, \texttt{ExoTraj} reduces peak muscle activation by approximately $41.3\%$, $10.9\%$, and $32.9\%$ for the three terrain conditions;
2) Compared with the No exo condition, \texttt{ExoTraj} achieves peak muscle activation reductions of $23.8\%$, $11.0\%$, and $-1.2\%$ for the three respective terrains;
3) Relative to the Assist off condition, the integrated muscle activation over the gait cycle is reduced by $31.9\%$, $-5.6\%$, and $30.8\%$ across the three terrains, respectively;
Under Assist on condition, compared with No exo, muscle-related metrics are reduced in certain scenarios, primarily due to the excessive weight of the entire system.
Furthermore, the magnitude of torque peaks on sloped and stair-climbing terrains is higher than that on level ground, and the gait phase corresponding to the peak muscle activation is close to those of the torque peaks, which validates that the generated torques are well aligned with the user’s intrinsic locomotion trend.

\begin{figure*}
	\centering
	\includegraphics[width=1\textwidth]{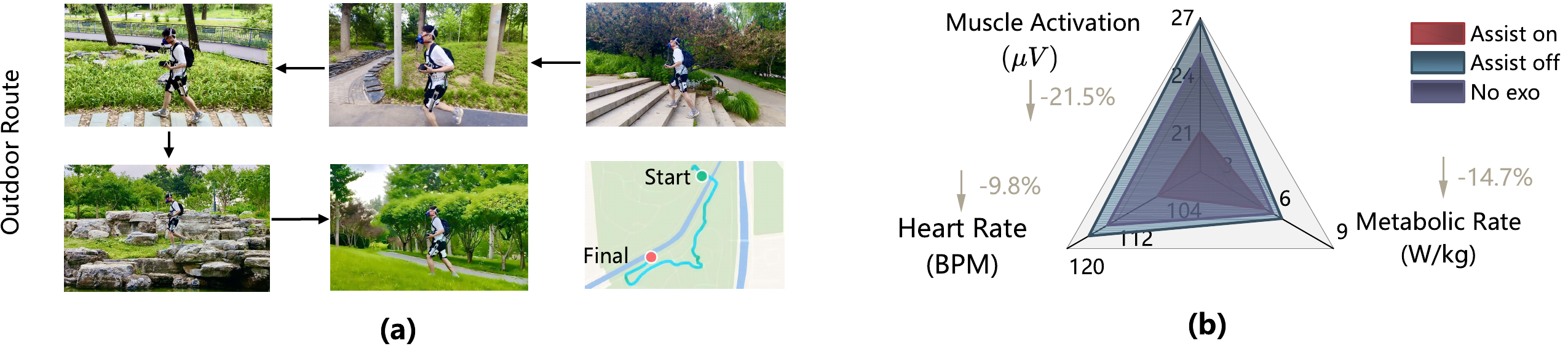}
    \captionsetup{justification=justified, singlelinecheck=false}
	\caption{ The comparative results of metabolic rate, average muscle activation, and heart rate after walking along a prescribed route on outdoor complex terrain under the three assistance conditions.}. 
	\label{fig10_1}
\end{figure*}

\begin{figure*}
	\centering
	\includegraphics[width=1\textwidth]{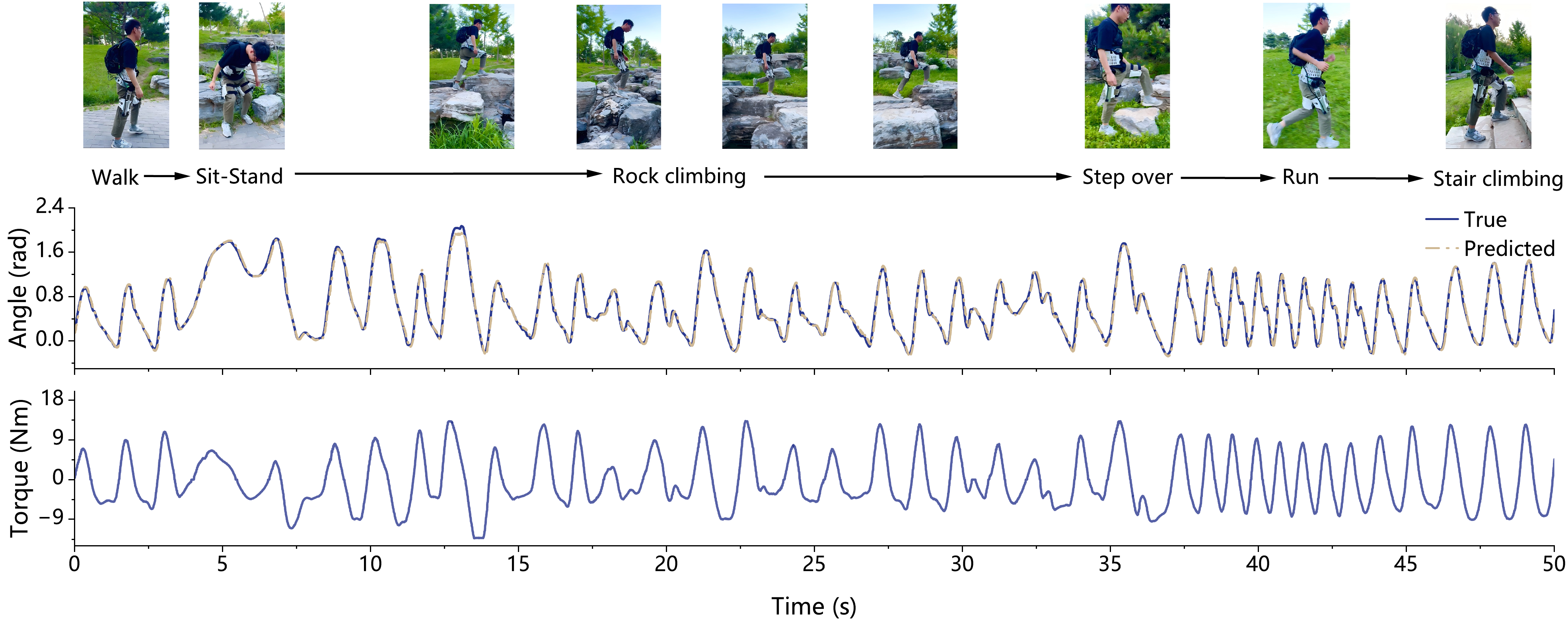}
    \captionsetup{justification=justified, singlelinecheck=false}
	\caption{ Results of real-time trajectory prediction and optimized torque profiles for cross-subject adaptive locomotion in complex outdoor field scenarios. The dashed lines represent the predicted trajectories, while the solid lines denote the actual joint trajectories.}. 
	\label{fig12}
\end{figure*}

\subsubsection{Metabolic and Heart Rate Reduction:} In this part, the subjects' metabolic rates are measured under two distinct scenarios. The first scenario consisted of relatively fixed repetitive movements, including level walking ($1.25$ m/s), slope walking (\ang{5}, $1.25$ m/s), and stair climbing ($70$ steps per minute). In this phase, each experimental trial lasts about $5$ minutes, and the average values of oxygen consumption and carbon dioxide production over the final $2$ minutes are used to determine the metabolic rate. 
Similarly, heart rate is determined using the average value recorded during the final $2$ minutes of each trial.
The second scenario involved a complex outdoor environment along the route, where subjects are instructed to adjust their walking speed to precisely match that of a designated leader waking speed in each of the distinct task conditions. The test route covers a range of complex outdoor terrains, including steep slopes, rocky terrain, and grassy surfaces. The total time to complete the entire experimental course is approximately $6$ minutes. Given that the entire experimental course comprises multiple distinct terrain segments, all performance metrics presented herein are the average values over the complete traversal.

Fig. \ref{fig10} shows the comparative results of the metabolic rate and heart rate average over three subjects recorded under three assistance conditions (Assist on, Assist off, and No exo) across three types of terrain. 
The results demonstrate that the \texttt{ExoTraj} policy significantly reduces metabolic rate and heart rate across all three assistance conditions: 1) Relative to the Assist off condition, the metabolic rate reductions range from $11.5\%$ during level walking to $24.4\%$ during stair climbing, and the heart rate reduction range from $1.7\%$ for level ground walking to $17.5\%$ for stair climbing. 2) 
Compared with the No exo condition, the metabolic reductions and heart rate range from $0.8\%$ to $18.2\%$ and from $-0.9\%$ to $11.0\%$. Stair climbing results in the highest energy expenditure, and the most significant reductions in metabolic rate and heart rate are observed, confirming the efficacy of the exoskeleton-assistance policy. Additional assistance tasks, such as ramp (\ang{10}, $1.25$ m/s), run ($1.8$ m/s), and sit-stand, comparison results can be found in Appendix \ref{other subject and sce}.

Fig. \ref{fig10_1} compares the metabolic rate, average muscle activation level, and average heart rate averaged over two subjects under different assistance conditions on complex outdoor terrain. The terrain composition and duration of the entire experimental route are as follows: stair ascent and descent (about $1$ minute), running (about $1$ minute), brisk walking (about $1$ minute), rock climbing (about $1$ minute), and walking on uneven grass terrain (about $2$ minutes). 
Fig. \ref{fig10_1} shows that \texttt{ExoTraj} effectively reduces key physiological metrics during locomotion: 1) Compared with the Assist off condition, \texttt{ExoTraj} yields relative reductions of $14.7\%$ in metabolic cost, $9.8\%$ in heart rate, and $21.5\%$ in muscle activation;
Relative to the No-exo baseline, ExoTraj achieves equivalent relative decreases: $8.1\%$ for metabolic cost, $7.4\%$ for heart rate, and $16.2\%$ for muscle activation.
The above experiments further demonstrate that the \texttt{ExoTraj} can effectively reduce energy expenditure in complex outdoor environments.

Fig. \ref{fig12} presents the trajectory and torque curve of the proposed \texttt{ExoTraj} method during real-world in-situ assistance operation in complex unstructured outdoor terrains. Tasks include walking and running on lawns, sit-to-stand transitions, rock climbing, obstacle crossing, and stair climbing, among others. As illustrated in Fig. \ref{fig12}, integrating trajectory prediction with torque optimization based on predicted trajectories enables the proposed framework not only to predict future locomotion trajectories accurately but also to achieve robust, adaptive performance in complex, unstructured locomotion tasks. As locomotion speed increases, the peak magnitudes of the generated torque profiles increase correspondingly. Moreover, the system exhibits robust adaptability to rapid changes in walking velocity. This demonstrates the high adaptability of \texttt{ExoTraj} to complex outdoor environments.
Additional results of predicted trajectories and assistance torques across diverse scenarios and high-dynamic movements (not seen in the training dataset) are provided in Appendix \ref{other subject and sce}.

	\section{Discussion and Conclusion}\label{conclusion}
This section provides further discussion on the necessity of trajectory prediction and exoskeleton robot assistance in high-dynamic environments, presents the advantages and limitations of the proposed method, and outlines promising directions for future improvements. Finally, the part concludes with a comprehensive summary of the entire work.

\subsection{Discussion}
\subsubsection{Advantages of Trajectory Prediction:}
The proposed trajectory prediction strategy outperforms the conventional single-step prediction approach in two critical aspects: 1) it enables torque optimization to leverage long-term temporal context information; 2) it supports look-ahead prediction, which effectively mitigates the adverse impacts of inference latency in the real-time control loop. Specifically, relying solely on a single trajectory is insufficient to achieve effective optimization of the torque profiles, which compromises the adaptability and robustness to complex locomotion tasks. For example, as shown in Fig. \ref{fig7} (a), a large deviation between predicted and actual values may easily induce control fluctuations, thereby hindering effective assistance. 

Furthermore, when the inference time exceeds the preset control time step, the single-step prediction scheme suffers from inherent time-lag behavior, which directly degrades the assistance comfort during human-exoskeleton interaction. For example, as listed in Table \ref{performance}, the LSTM takes about $10$ ms for inference, which cannot satisfy the $5$ ms real-time control requirement and restricts the deployment of complex models for single-step prediction.
Moreover, most multi-modal data in existing exoskeleton systems are time-series, enabling fast inference.
Integrating linguistic information for natural human-exoskeleton interaction and visual information for complex terrain perception will significantly increase model inference latency. The trajectory prediction strategy can effectively mitigate this latency issue via pre-executed look-ahead prediction, which is a key advantage of trajectory prediction for future complex locomotion tasks.

\subsubsection{Assistance in High-dynamic Environments:}
For exoskeleton robots to enhance lower-limb walking ability and reduce energy consumption in outdoor environments, they must adapt to complex and dynamic conditions. Traditional rule-based methods struggle to meet these assistance requirements. Over the past decade, neural network approaches have trained parameter weights using large amounts of feature-label data pairs, thereby learning a complex, high-dimensional function that maps features to labels in complex scenarios. Consequently, high-quality human-robot interaction (HRI) data has become the cornerstone of effective adaptive assistance for exoskeleton robots. While human kinematic features such as joint positions and velocities can be readily acquired through wearable sensors, accurate kinetic labels, including joint torques and ground reaction forces, remain extremely challenging to obtain directly in unconstrained environments. 

To overcome this data acquisition bottleneck, two main research directions have been explored in the literature. The first approach calculates joint torques through inverse dynamics using data captured by optical motion capture systems \citep{molinaro2024estimating,molinaro2024task}. However, such systems require expensive and complex laboratory setups, making them impractical for deployment in complex outdoor terrains. The second approach generates synthetic torque data by training musculoskeletal models to perform human-like locomotion in simulation environments \citep{luo2024experiment}. Nevertheless, musculoskeletal models typically involve high-dimensional control spaces (often exceeding 100 dimensions), which makes it notoriously difficult to train them to reproduce natural human walking behaviors across diverse and complex terrains.

Consequently, existing approaches remain fundamentally limited in their ability to provide robust exoskeleton assistance across diverse, unstructured outdoor terrains. To address both the prohibitive data acquisition costs and the insufficient environmental adaptability of current methodologies, this work presents a hierarchical control framework. The high-level module is designed to predict locomotion information, specifically joint angles and angular velocities, which can be directly captured using low-cost wearable sensors with minimal deployment complexity. The low-level module then establishes a mapping between the predicted kinematic trajectories and the corresponding desired assistance torques using MPC. The proposed hierarchical framework can reduce the data acquisition burden and enhance the exoskeleton's adaptability to complex and dynamic outdoor environments. More discussions on exoskeleton assistance are provided in Appendix \ref{other discuss}.

\subsubsection{Discussion for Policy \texttt{ExoTraj}:} Human locomotion characteristics exhibit significant inter-subject variability. The proposed assistance policy \texttt{ExoTraj} proposes a fast flow matching method to model the probabilistic distribution of trajectories, thereby generating reference trajectories in real time with enhanced accuracy and generalization capability, as demonstrated in Table \ref{performance} and Fig. \ref{fig6}. Perception and control are inherently strongly coupled in exoskeleton systems. Building upon a dynamic model, the proposed assistance policy \texttt{ExoTraj} introduces a novel objective function for MPC, which generates smooth and comfortable assistance torque curves. This approach maintains robust control performance even in the presence of external disturbances, as illustrated in Table \ref{robustness} and Fig. \ref{fig7}. Compared with the assist off baseline, \texttt{ExoTraj} significantly reduces both the user's metabolic rate and muscle activation levels in outdoor environments.

Nevertheless, the proposed policy \texttt{ExoTraj} still has several limitations. 
The torque optimized by the MPC controller based on predicted trajectories still exhibits discrepancies with the ground truth joint torque curves. 
Although the torque optimized by \texttt{ExoTraj} generally agrees with human locomotion torques, noticeable discrepancies remain in some local details compared with the ground-truth torques.
In scenarios where ground truth joint torques are not directly measurable, neural network optimization based on user preferences provides a promising alternative. However, current preference-based optimization methods for exoskeletons have been predominantly limited to tuning low-dimensional control parameters, rather than optimizing complex high-dimensional neural networks, which remains a significant open challenge in the field. 
Additionally, the overall weight of the current exoskeleton prototype remains relatively high, which may limit its long-term outdoor use. Future work will focus on comprehensive system optimization to reduce weight and enhance portability. 
Another critical yet underinvestigated question in lower-limb exoskeleton control design pertains to establishing the appropriate scaling relationship between ground-truth hip joint moments and exoskeleton assistance torques when the former is experimentally measured. During the stance phase, a substantial portion of the inherent hip joint moments is dedicated to supporting body weight and maintaining postural equilibrium. A fundamental ambiguity remains regarding whether these load-bearing and balance-related torque components should be explicitly incorporated into the exoskeleton's assistance torque. 

\subsection{Conclusion}
This paper presents a novel lower-limb exoskeleton assistance policy, named \texttt{ExoTraj}, designed for complex unstructured outdoor environments. The proposed policy features low-cost data acquisition, requiring only that the user walk naturally while donning the exoskeleton. \texttt{ExoTraj} employs a hierarchical control framework, wherein the high-level controller leverages fast FM to enable real-time, high-precision prediction of joint trajectories; the low-level controller employs MPC to optimize torque, thereby delivering comfortable and robust assistance. 
Compared with conventional data-driven approaches, \texttt{ExoTraj} reduces trajectory prediction error by up to $14.0\%$ under cross-subject and complex terrain conditions. Relative to traditional control methods, its optimized torque achieves superior smoothness and enhanced robustness in highly dynamic scenarios. When applied in complex dynamic outdoor environments, \texttt{ExoTraj} reduces the metabolic rate, heart rate, and peak muscle activation by up to $24.4\%$, $19.5\%$, and $41.3\%$, respectively. \texttt{ExoTraj} establishes a generalizable framework for outdoor lower-limb exoskeleton assistance, and further advances the translation of exoskeleton robots from laboratory research to real-world daily applications. 


    
    \begin{funding}
    This work was funded by the National Natural Science Foundation of China under (Grant 62473365, Grant U22A2056, and Grant 62373013), and the Beijing Natural Science Foundation under (Grant L232021 and L242101)        
    \end{funding}
    
    \begin{dci}
    The authors declared no potential conflicts of interest with respect to the research, authorship, and/or publication of this article.
    \end{dci}

    \newpage
    \clearpage
\newpage
\appendix  
\appendixpage  
\addappheadtotoc  

\setcounter{equation}{0}
\renewcommand{\theequation}{A.\arabic{equation}}
\setcounter{figure}{0}
\renewcommand{\thefigure}{A.\arabic{figure}}
\setcounter{table}{0}
\renewcommand{\thetable}{A.\arabic{table}}
\setcounter{section}{0}
\renewcommand{\thesection}{A.\arabic{section}}

\section{Notations}\label{notation}
The primary mathematical notations employed in this paper are summarized in Table \ref{tab_notation}. The notations are divided into two parts: one concerns trajectory generation (i.e., the proposed fast flow matching algorithm), and the other focuses on trajectory optimization, which is related to model predictive control.
\begin{table}[htbp]
	\normalsize
	\caption{Main symbols and notations used in this work.}
	\label{tab_notation}
	\centering
\begin{tabular}{cp{5.8cm}}
\toprule[1pt] 
{Notation}& {Description}\\
\midrule
\multicolumn{2}{c}{\cellcolor{gray!10}\textbf{Fast Flow Matching}}\\
$v_t$& Velocity vector\\
$\psi_t$& Flow path \\
$\mathbf{x}$& Input noise \\
$\mathbf{o}$&Observations in the past $T_o$ steps\\
$\mathbf{y}$& Trajectory information of hip joint\\
$\mathbf{a}$& Motor torque of future $N$ steps\\
$T_o$& Time steps of the observation\\
$T_p$& Time steps of the prediction trajectory\\
$L$& Usage steps of predicted trajectory\\
\midrule
\multicolumn{2}{c}{\cellcolor{gray!10}\textbf{Model Predictive Control}}\\
$N$& Optimization horizon of MPC\\
$\mathbf{M}$& Inertial matrix of dynamic model\\
$\mathbf{C}$& Velocity matrix of dynamic model\\
$\mathbf{C}$& Gravity matrix of dynamic model\\
$\mathbf{T}_{int}$& Interaction torque\\
$\mathbf{Q}$& Tracking error matrix\\
$\mathbf{R}$& Torque constraint matrix\\
$\mathbf{P}$& Penalization weight matrix\\
$\mathbf{O}$& Terminal weight matrix\\
\bottomrule[1pt]
\end{tabular}
\end{table}

\begin{figure*}[b]
	\centering
	\includegraphics[width=0.98\textwidth]{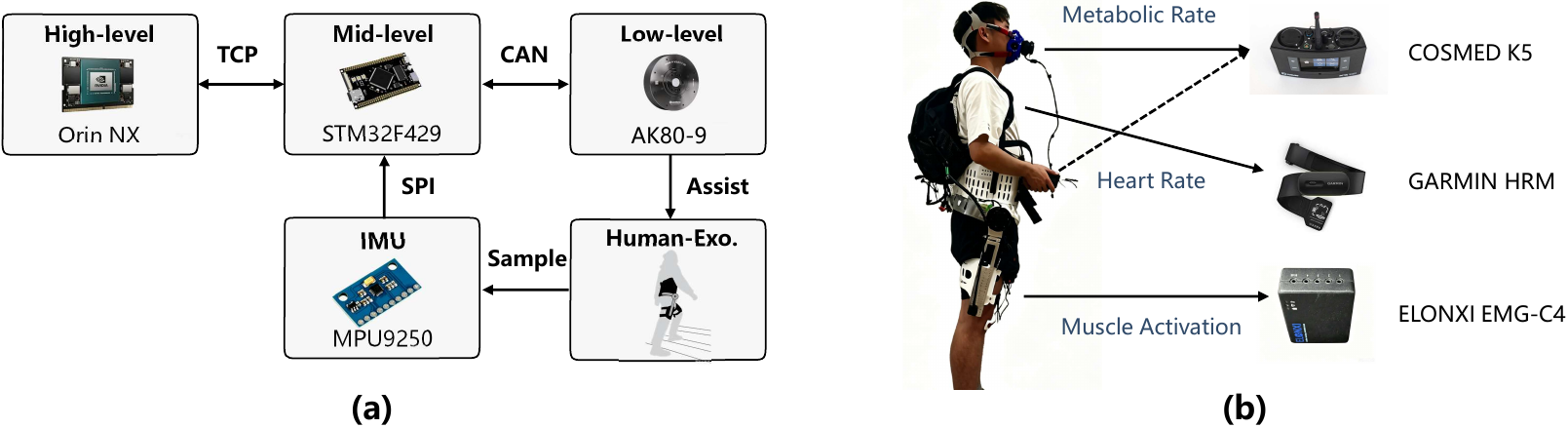}
    \captionsetup{justification=justified, singlelinecheck=false}
	\caption{(a) shows the system interconnection and communication architecture: TCP for data exchange between the Jetson Orin NX and STM32, CAN bus for STM32-drive motor communication, and SPI for IMU data acquisition by the STM32. (b) presents the relevant equipment used for measuring metabolic rate, muscle activation, and heart rate.}. 
	\label{figa}
\end{figure*}

\section{Other Experimental Details}
This section provides comprehensive details concerning the exoskeleton system, real-time inference data flow, experimental participants, and data acquisition procedures.
\subsection{Details of System Description} \label{app_composition}
Fig. \ref{figa} (a) illustrates the overall architecture of the proposed system design. The exoskeleton's electronic system integrates perception, computation, and actuation into a cohesive unit. The high-level control, powered by an NVIDIA Jetson Orin NX, performs real-time trajectory prediction and torque optimization. A centralized STM32 board handles data acquisition, gathering motor data via CAN and IMU data via SPI, while facilitating bidirectional Ethernet communication with Jetson. Low-level motor driving is delegated to the AK80-9 T-Motor internal controllers, which also provide real-time current, velocity, and position feedback. The sensor data acquisition and TCP data transmission are strictly regulated with a $5$ ms cycle, while the low-level control operates at a $1$ ms period. 
Fig. \ref{figa} (b) presents a schematic diagram of the equipment placement for measuring the physiological metrics of metabolic rate, muscle activation, and heart rate. The three types of sensory data are synchronously acquired, and precise data alignment is performed based on their corresponding recorded timestamps.

\begin{figure*}
	\centering
	\includegraphics[width=0.96\textwidth]{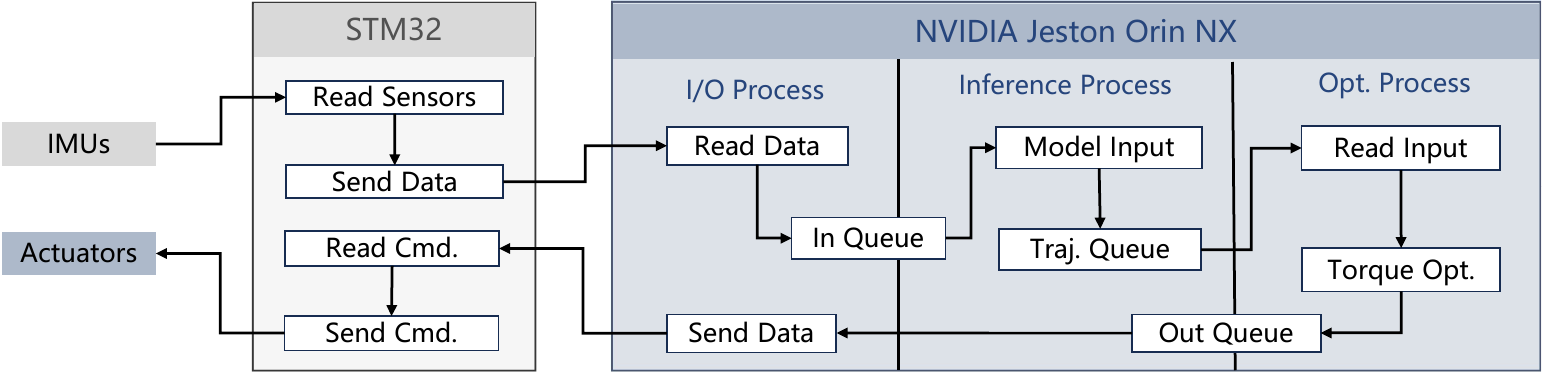}
    \captionsetup{justification=justified, singlelinecheck=false}
	\caption{Data flow diagram of the exoskeleton robot system. Sensor data is read by the STM32 via SPI and CAN protocols, then transmitted to the Orin NX via TCP over Ethernet. Three threads run in parallel on the edge device: the I/O thread reads sensor data in real time every 5 ms; new trajectories are buffered when the inference trigger is met. The MPC optimizes torque in real time based on the next N-step trajectory and sends it to the STM32 every 5 ms to achieve real-time inference control.}. 
	\label{figa1}
\end{figure*}

In complex scenarios, real-time performance is critical for the control of exoskeleton robots. The high-level controller consists of two modules: trajectory prediction and torque optimization. It is required to ensure that the upper-level decision-making process is completed within $5$ ms before being transmitted to the low-level controller, where the time consumption for data reading and transmission must also be taken into account. The edge computing device employed is the Jetson Orin NX, which supports both GPU acceleration and general-purpose CPU computation.
To ensure real-time computation, a parallel computing architecture is adopted for the aforementioned trajectory prediction, torque optimization, and data read/write operations. 
Fig. \ref{figa1} presents a detailed data flow diagram of the exoskeleton robot system. The main processor streams data from the IMUs and actuator encoders to the machine learning coprocessor. The coprocessor runs three concurrent processes: an I/O handler for data communication and preprocessing, a dedicated engine for minimal-latency model inference, and a module for optimizing model-predicted control torque, which is then returned to the main processor. To accelerate the inference speed of Fast FM, the pre-trained model is first converted to the ONNX format and further optimized into a TensorRT engine to maximize inference throughput. To guarantee both the stability and computational efficiency of MPC optimization, the acados library in Python is utilized for numerical solution in this work. All I/O and queue operations between the processors employ non-blocking logic, ensuring the exoskeleton's control loop robustly maintains a frequency of $200$ Hz.

\begin{table}
	\normalsize
	\caption{Physical information of each subject. The first eight are for offline multi-modal data acquisition, and the latter four are for cross-subject validation experiments.}
	\label{tab_subject}
	\centering
	\resizebox{7.8cm}{!}{
\begin{tabular}{ccccc}
\toprule[1pt] 
Subject ID& Age & Gender &Height (m)& Mass (kg)\\

\midrule
AB01& 25 & M &1.78& 81.0\\
AB02& 24 & M &1.86& 58.4\\
AB03& 24 & M &1.75& 75.2\\
AB04& 22 & M &1.75& 64.0\\
AB05& 22 & M &1.80& 80.5\\
AB06& 23 & M &1.75& 76.6\\
AB07& 24 & F &1.65& 52.1\\
AB08& 26 & F &1.68& 53.0\\
AB09& 24 & M &1.83& 78.6\\
AB10& 21 & M &1.77& 85.3\\
AB11& 20 & M &1.70& 66.5\\
AB12& 26 & M &1.67& 65.5\\
\bottomrule[1pt]
\end{tabular}
}
\end{table}

\subsection{Details of the Collected Dataset} \label{dataset}
We recruited participants with diverse heights and body weights to collect multimodal data during walking on complex outdoor terrains. Detailed participant information is provided in Table \ref{tab_subject} ($10$ males, $2$ females; age: $23.4 \pm 3.5$ years; height: $174.9 \pm 0.4$ cm; weight: $69.7$ kg). Before the experiment, participants were informed of the experimental procedure, details, and precautions. All provided written informed consent (approval number: IA-2502-020403). Each data acquisition session lasted approximately $15$ minutes, and two sessions were conducted for each participant. The first eight subjects were instructed to walk freely along the designated path at a self-selected speed under the guidance of a trained experimenter. The walking terrain included complex sloped surfaces, irregular lawns, and inclined ramps. 
The exoskeleton operated in zero-torque assistance mode. During walking, we collected data from 3 IMUs providing 18-dimensional information (position and velocity) and 2 motors providing 4-dimensional information (joint angle and angular velocity). Among these, the 18-dimensional IMU data and 2-dimensional joint angle data were used as input features, while the 2-dimensional joint angle and 2-dimensional angular velocity data served as the prediction labels. The data sampling frequency was set to 1 kHz. The raw angular velocity signals were processed using a low-pass filter to suppress noise. Finally, the training dataset was downsampled to 200 Hz. In addition, owing to variations in joint positions across different participants, calibration of the initial motor angles was performed before data collection. The initial reference was defined as the static standing posture of the participant while wearing the exoskeleton.

\section{Other Experimental Results}\label{other_results}
This section presents a comprehensive analysis of the model-related hyperparameters, including the effective trajectory length $L$, the number of observation steps $T_o$ and prediction steps $T_p$, the number of iteration steps for FM, and the key parameters of the MPC module. Subsequently, we compare the MPC-optimized torque with the measured actual torque collected from public datasets, and analyze the discrepancies between the two. Furthermore, this section presents the experimental results for additional scenarios and subjects.

\subsection{Analysis of Usage Steps \texorpdfstring{$L$}{L}} \label{ablation_L}
The performance of model predictive control (MPC) for lower-limb exoskeletons is critically determined by both the accuracy and timeliness of predicted joint trajectories. While a longer prediction horizon length ($L$) theoretically yields smoother control actions and improved anticipation of human motion intent, it inevitably introduces accumulated prediction errors and increased computational latency.
To systematically investigate the impact of $L$ on online trajectory prediction performance, we conduct cross-subject treadmill walking experiments at speeds ranging from $0$ m/s to $1.6$ m/s. Five distinct values of the effective prediction horizon length $L$ are selected for evaluation: $16, 20, 40, 80,$ and $100$. since the MPC prediction horizon $N$ is set to $15$, the number of steps used cannot be less than $15$. Trajectory prediction accuracy is quantified using the ADE (See Eq. \eqref{pred_acc}), which measures the discrepancy between predicted and ground-truth trajectories.
Please note that in this experiment, the exoskeleton provides torque assistance.

\begin{figure}
	\centering
	\includegraphics[width=0.48\textwidth]{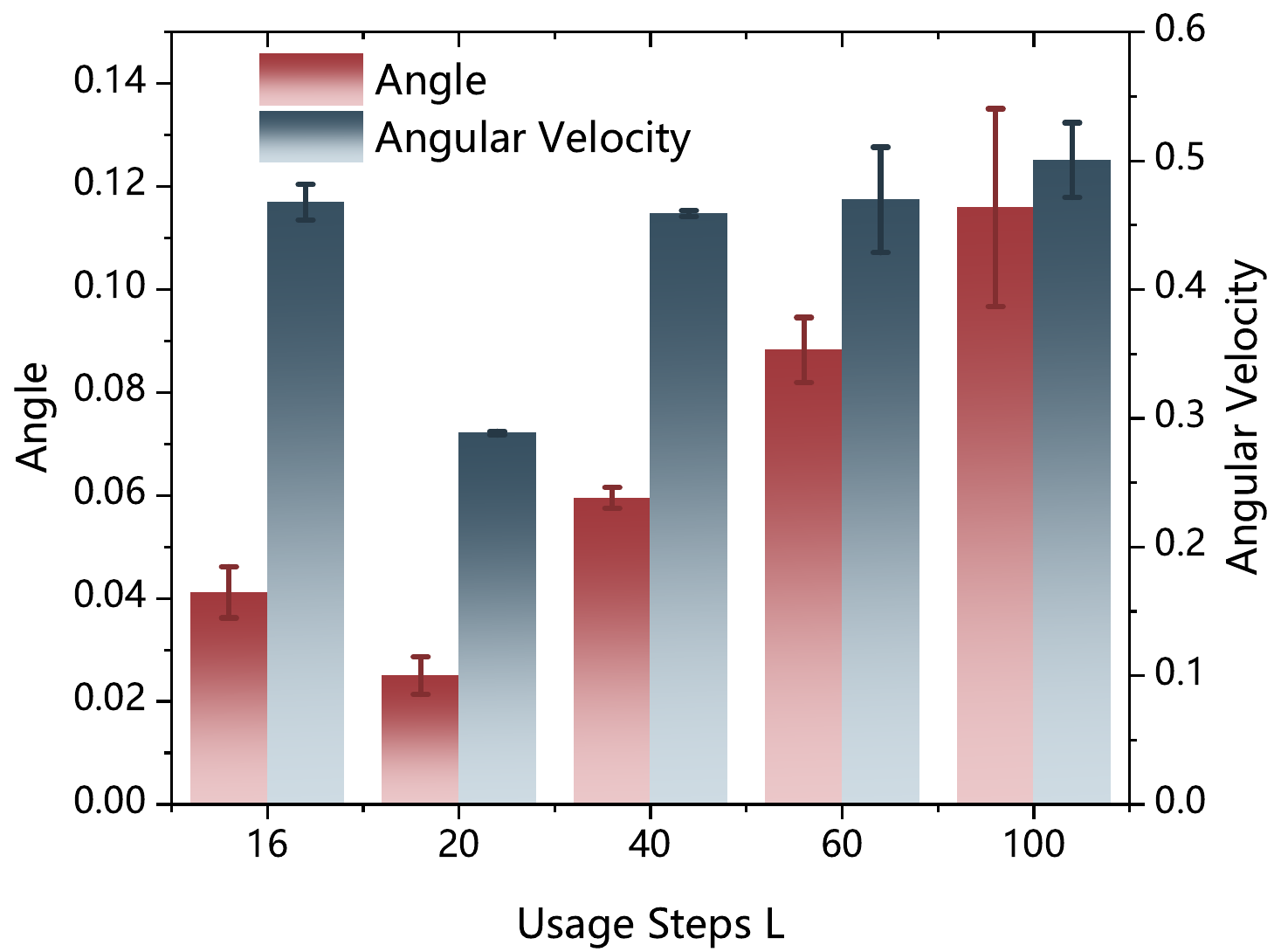}
    \captionsetup{justification=justified, singlelinecheck=false}
	\caption{Variation of online trajectory (angle and angle velocity) prediction error with different usage steps $L$.}. 
	\label{figa2}
\end{figure}

Fig. \ref{figa2} presents the effect of different lengths of $L$ on the prediction errors of angle and angular velocity under exoskeleton assistance.
The results demonstrate that during actual exoskeleton assistance, the error exhibits an upward trend as $L$ increases, which is consistent with the phenomenon that the longer the network prediction horizon, the larger the prediction error. However, the error is relatively large at $L=16$ (one inference per torque optimization). This may be due to continuous trajectory updates causing discontinuous torque optimization and non-optimal values, which impair assistance performance, affect user gait, and reduce prediction accuracy. 
Additionally, when the model incurs non-negligible inference latency (e.g., $50$ ms, which corresponds to a $10$-step lag), the number of steps utilized must be increased accordingly to mitigate the adverse effects introduced by such computational delay. Specifically, for an MPC prediction horizon $N$ of $15$ steps under a $10$-step lag condition, the minimum required number of usage steps is $25$. For the implementation in this work, the model exhibits an average inference time of $6$ ms, which necessitates $1$-step look-ahead inference. As a result, the final usage step count $L$ is configured to $20$.

\begin{figure}
	\centering
	\includegraphics[width=0.46\textwidth]{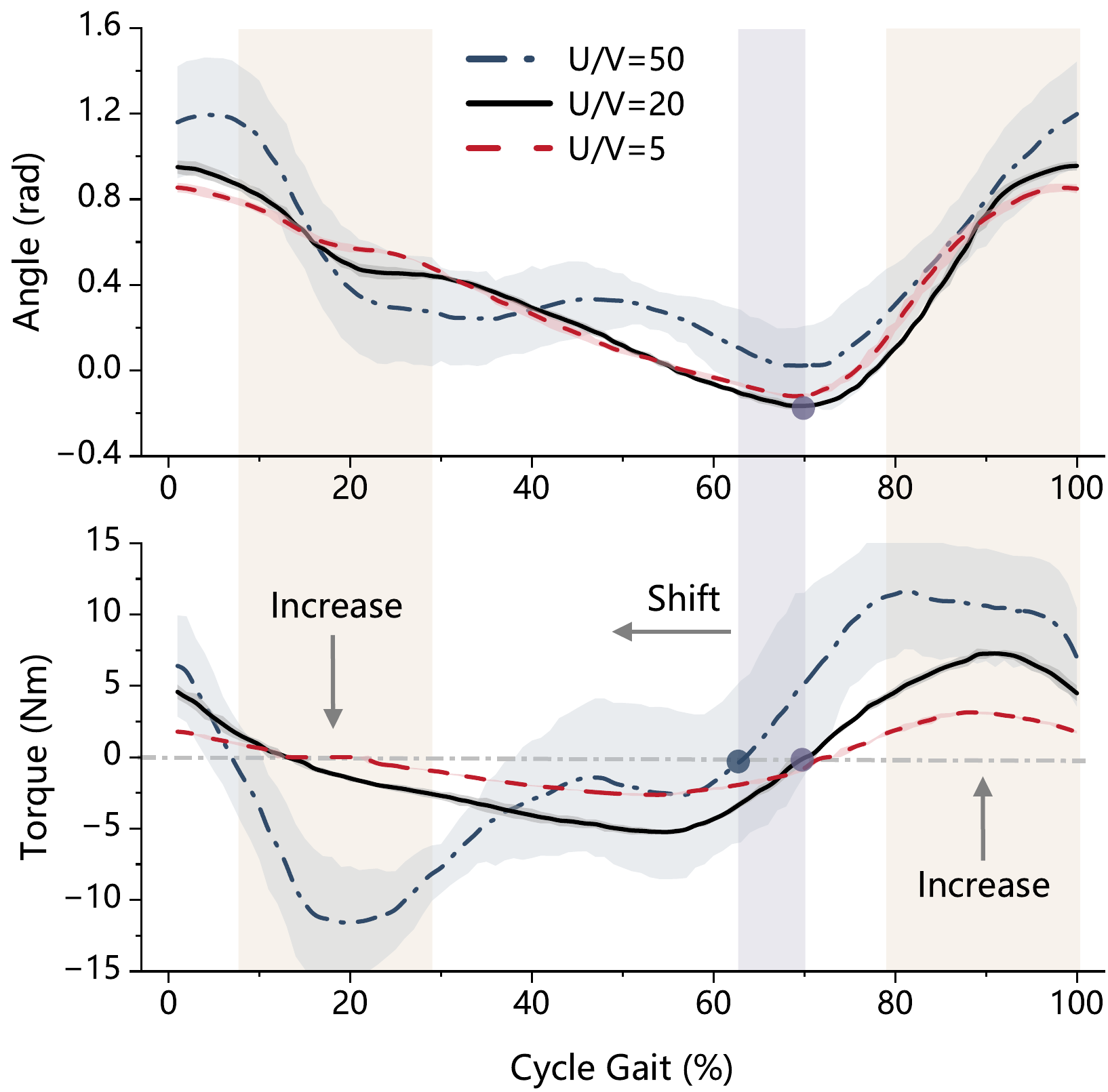}
    \captionsetup{justification=justified, singlelinecheck=false}
	\caption{Comparison results of real-time trajectory prediction and torque optimization under different \(\mathbf{U}/\mathbf{V}\) ratios during treadmill walking at $1.0$ m/s across subjects.}. 
	\label{figa3}
\end{figure}

\subsection{Analysis of MPC Parameters}
This subsection presents a detailed analysis of the influence exerted by the parameters \(\mathbf{U}\) and \(\mathbf{V}\) in the model predictive control (MPC) objective function (Eq. \eqref{eq_20}) on the resulting optimized torque profiles. Here, \(\mathbf{U}\) and \(\mathbf{V}\) denote the penalty terms for trajectory tracking error and torque regularization, respectively. Specifically, the trajectory tracking penalty term \(\mathbf{U}\) is determined by the error weighting matrix \(\mathbf{Q}\) and the terminal weighting matrix \(\mathbf{O}\), while the torque regularization penalty term \(\mathbf{V}\) is governed by the torque regularization matrix \(\mathbf{R}\) and the control penalty matrix \(\mathbf{P}\).
Since the final form of the MPC objective function comprises these two penalty terms exclusively, analyzing their relative ratio is sufficient to characterize the impact of the aforementioned matrix parameters on the optimized torque profiles. Given that the error weighting matrix \(\mathbf{Q}\) consists of two components corresponding to the trajectory angular error and angular velocity error, we fix the weighting ratio of angular error to angular velocity error at \(100:1\) throughout our analysis. Consequently, the ratio of penalty term \(\mathbf{U}\) to \(\mathbf{V}\) is defined as the ratio of the angular error coefficient to the magnitude of \(\mathbf{V}\).

\begin{table}
	\normalsize
	\caption{Performance metrics of optimized torque profiles for exoskeleton online assistance under different \(\mathbf{U}/\mathbf{V}\) ratio.}
	\label{taba2}
	\centering
	\resizebox{8.5cm}{!}{
		\begin{tabular}{c|ccc}  
			\toprule[1pt] 
			$\mathbf{U}/\mathbf{V}$ \textbf{Ratios}&$50:1$&$20:1$& $5:1$ \\
			\midrule[1pt]  
            Smooth $S_\text{rms}$ &$16.7\pm1.30$ & $8.20\pm 0.16$&$\mathbf{3.72\pm0.05}$\\
			Phase $R_\text{shift}$ &$0.05\pm0.04$ &$\mathbf{0.01\pm0.01}$ &$0.02\pm0.01$\\  
            Comfort $C_\text{mpc}$ &$0.90\pm0.02$ &$0.96\pm0.00$ &$\mathbf{0.97\pm0.00}$\\
			\bottomrule[1pt]  
		\end{tabular}
	}
\end{table}

Fig. \ref{figa3} presents the comparison of the predicted trajectories and optimized torque profiles during the online assistance process, when a human subject walks on the treadmill under three different penalty ratios of $\mathbf{U}$ to $\mathbf{V}$: $5:1$, $20:1$, and $50:1$. 
As the \(\mathbf{U}/\mathbf{V}\) ratio decreases, corresponding to a reduction in the degree of torque regularization, both the peak torque values and the fluctuations in the torque profiles increase, and gait consistency deteriorates. Conversely, as the \(\mathbf{U}/\mathbf{V}\) ratio increases, corresponding to enhanced torque regularization, the optimized torque profiles become progressively smoother; however, this improvement is achieved at the expense of losing fine-grained local details.
From a mathematical perspective, an increase in the torque regularization coefficient directly corresponds to a larger torque smoothing coefficient as defined in Eq. \eqref{eq_18}, which in turn yields smoother torque profiles. In contrast, a smaller torque penalty coefficient allows the torque profiles to retain richer local details but compromises the comfort of the assistance provided.

Table \ref{taba2} summarizes three key performance metrics of the optimized torque profiles under different \(\mathbf{U}/\mathbf{V}\) ratio configurations: smoothness $S_\text{rms}$, phase consistency $R_\text{shift}$, and comfort metric $C_\text{mpc}$ defined in Eq. \eqref{eq_22}-Eq. \eqref{comfort}. 
Table \ref{taba2} confirms that decreasing the \(\mathbf{U}/\mathbf{V}\) ratio improves both the comfort index and smoothness index. However, these improvements come at the cost of reduced torque optimization authority, substantial loss of fine-grained local details, and consequently diminished assistance efficiency. Conversely, increasing the \(\mathbf{U}/\mathbf{V}\) ratio leads to more pronounced fluctuations in the torque profiles, degraded smoothness, and increased gait inconsistency, which negatively impacts user comfort. These quantitative findings align closely with the qualitative observations presented in Fig. \ref{figa3}. To achieve an optimal balance among assistance comfort, efficiency, and preservation of local details, this work selects the intermediate value of $20$ as the final ratio coefficient for the MPC objective function.

\begin{figure}
	\centering
	\includegraphics[width=0.45\textwidth]{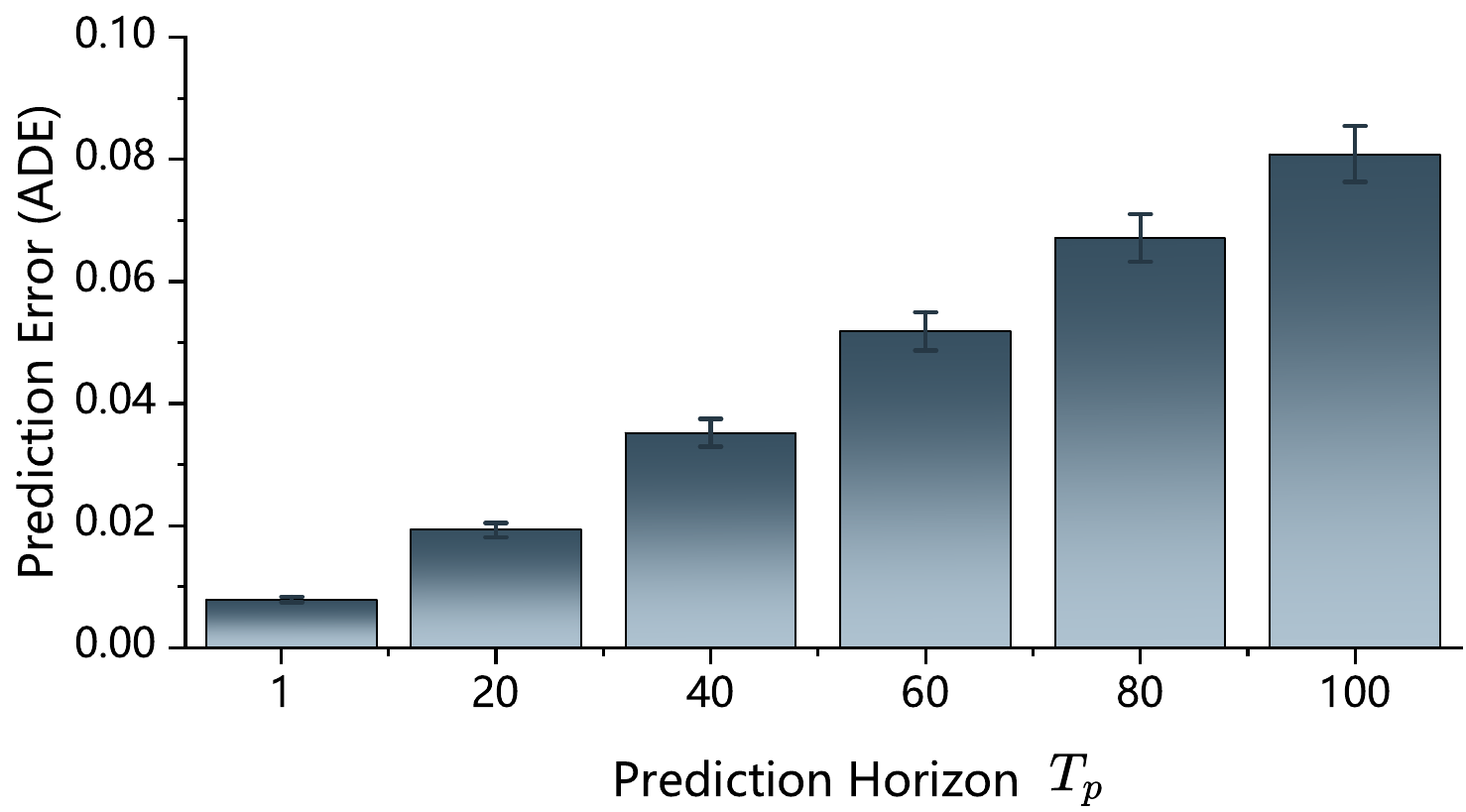}
    \captionsetup{justification=justified, singlelinecheck=false}
	\caption{Influence of prediction horizon \(T_p\) on real-time trajectory prediction error in offline phase.}. 
	\label{figa5}
\end{figure}

\subsection{Analysis of Observation Steps \texorpdfstring{$T_o$}{To} and Prediction Steps \texorpdfstring{$T_p$}{Tp}}
Intuitively, the prediction error increases as the prediction horizon $T_p$ grows longer. Fig. \ref{figa5} illustrates the impact of different prediction horizons on trajectory accuracy during the offline phase, which is consistent with the aforementioned observation. The MPC optimization requires a prediction horizon of no less than $15$ steps. Fig. \ref{figa2} demonstrates that a trajectory length of approximately $20$ yields satisfactory trajectory prediction accuracy during the real-time assistance process. In this paper, we set the prediction horizon to $50$, but only utilize the first $20$ steps for actual implementation. For the observation step $T_o$, an excessively short sequence contains limited information and thus leads to inaccurate trajectory prediction. Conversely, an overly long sequence is not suitable for highly dynamic scenarios. Due to drastic motion variations, a large time window fails to accurately predict the user’s intention at the next moment. Table \ref{taba3} shows the influence of observation steps $T_o$ for trajectory prediction accuracy in the offline dataset. The results show that prediction accuracy improves as the number of observation steps increases. The accuracy at $250$ steps is nearly identical to that at $500$ steps. 
This indicates that as the prediction horizon \(T_o\) increases, no significant improvement in prediction performance is observed beyond $250$ time steps. Furthermore, excessively long prediction horizons are unsuitable for highly dynamic scenarios.
To maintain adaptability to dynamic scenarios, we set the observation steps to $250$ in this work.

\begin{table}
	\normalsize
	\caption{Comparison results for different observation steps $T_o$ and the iteration steps (NEF) of FM.}
	\label{taba3}
	\centering
	\resizebox{8.4cm}{!}{
		\begin{tabular}{ccc|cccc}  
			\toprule[1pt] 
            \multicolumn{3}{c|}{\cellcolor{gray!20}\textbf{Observation Steps} $T_o$} &\multicolumn{4}{c}{\cellcolor{gray!20}\textbf{Iteration Steps (NEF)}}\\
			Num. &ADE&$R^2(\%)$ &Num.&ADE&$R^2(\%)$&Time\\
			\midrule[1pt]  
			$50$ & $0.0259$ & $99.88$& $1$ &${0.0136}$ & $99.76$ & $4.26$\\
			$250$ & $0.0044$ & $99.94$ & $3$&$\mathbf{0.0123}$ & $99.87$ & $12.70$\\ 
            $500$ & $\mathbf{0.0043}$ & $\mathbf{99.94}$ & $10$&${0.0149}$ & $99.79$ & $41.37$\\ 
			\bottomrule[1pt]  
		\end{tabular}
	}
\end{table}

\subsection{Analysis of NEF for FM}
This subsection analyzes the influence of the iteration steps of FM (NEF) on trajectory prediction performance. We analyze the offline cross-subject trajectory prediction error and inference latency under four settings with \(\text{NEF}=1, 3, 10\). It should be noted that all recorded inference times are measured on a workstation equipped with the RTX 4090 GPU, rather than edge computing devices. Table \ref{taba3} presents the corresponding results of different iteration steps for FM methods in the offline dataset. It can be concluded that, within the transformer encoder-decoder framework, higher iteration steps do not necessarily yield higher trajectory prediction accuracy. FM presents poor performance for one iteration. While three iterations enhance prediction accuracy, further increasing to ten iterations leads to accuracy degradation. This is mainly due to the low time utilization of the Transformer framework. First, it is only used in the encoding phase, but not in the decoding phase. Second, error accumulation exists: during the $[0, 1]$ time interval, the network fails to learn well in some time segments, which degrades the prediction accuracy and further impairs the final inference accuracy. The proposed fast FM method requires no iterative process and generates prediction results directly in one step.

\subsection{Comparison between MPC-optimized and Ground-Truth Torques}\label{torque_comparison}
\begin{figure*}
	\centering
	\includegraphics[width=1\textwidth]{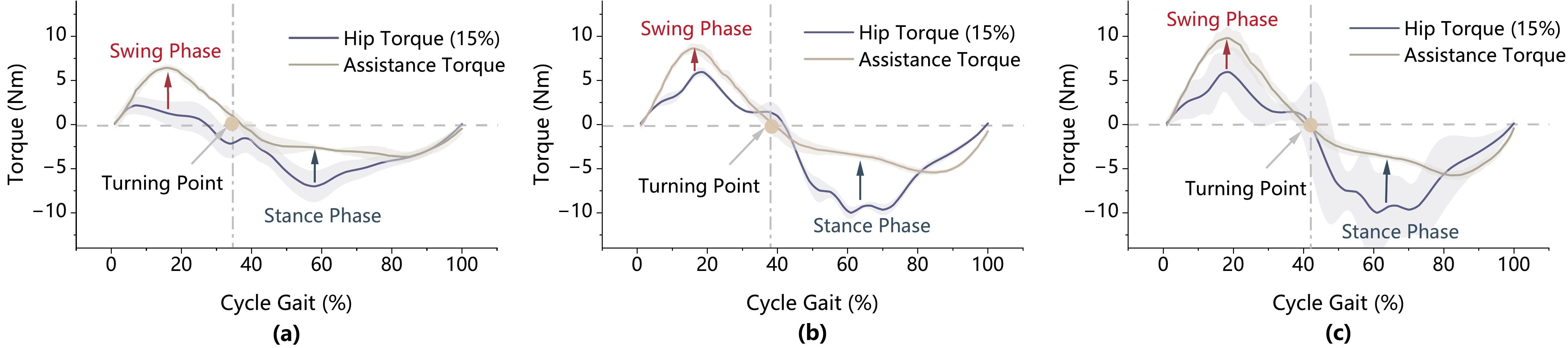}
    \captionsetup{justification=justified, singlelinecheck=false}
	\caption{Comparison between torques optimized by the \texttt{ExoTraj} policy and hip joint ground-truth torques (scaled by $0.15$). Subfigures (a), (b), and (c) present the torque comparisons at treadmill speeds of $0.6$, $1$, and $1.4$ m/s, respectively.}. 
	\label{figa_torque_comparison}
\end{figure*}

This part aims to quantitatively compare the correspondence between MPC-optimized assistance torques and human physiological joint torques. Given the inherent limitations of motion capture systems, which are restricted to controlled laboratory environments, we conduct this comparative analysis using publicly available biomechanical datasets \citep{camargo2021comprehensive}. Specifically, we select data from subjects in the public dataset whose anthropometric characteristics (body weight and height) closely match those of our cross-subject experiment participants. We extract joint torque data for treadmill walking at speeds of $0.6$, $1.0$, and $1.4$ m/s to compare with the MPC-optimized torques. Since the exoskeleton is designed to provide approximately $15$\% of the human physiological joint torque \citep{molinaro2024task}, we scale the ground-truth torque values from the dataset by a factor of $0.15$ to ensure a meaningful comparison. Fig. \ref{figa_torque_comparison} compares the MPC-optimized torques with the scaled hip joint ground-truth torques. 
It shows that human ground-truth hip joint torque curves exhibit significantly higher peak torques during the stance phase, whereas the swing phase is characterized by relatively low torque magnitudes. This is attributed to the fact that the stance phase must support the entire body weight, resulting in substantially higher torque demands.

\begin{figure*}
	\centering
	\includegraphics[width=1\textwidth]{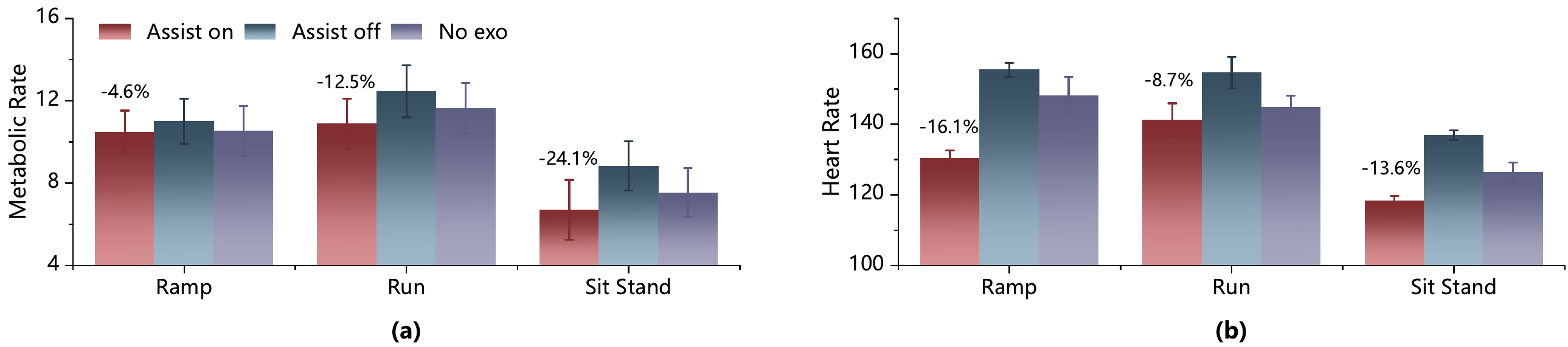}
    \captionsetup{justification=justified, singlelinecheck=false}
	\caption{The comparative results of metabolic rate (a) and heart rate (b) across different assistance modes during three task types: walking at a 10-degree ramp, running, and sit-to-stand transitions.}. 
	\label{figa7}
\end{figure*}

When humans walk, the thigh moves at high angular velocities during the swing phase, and hip joint torque is mainly used to flex the hip and lift the leg. During the stance phase, if we exclude the torque needed for body weight support and postural balance (produced primarily by the gluteus maximus and hamstrings through hip extension), the peak hip torque in this phase drops significantly.
For exoskeleton-assisted walking, the motor only provides hip extension torque. Since the exoskeleton is strapped to the user’s thigh and applies force directly to this segment, it cannot effectively support body weight or maintain postural balance. Therefore, in actual assistance scenarios, the stance phase does not require a high peak torque output from the exoskeleton.
During the swing phase, to reduce thigh muscle effort, the actuator torque should be increased accordingly, as faster leg lifting requires greater assistive torque. The \texttt{ExoTraj} policy optimizes the torque profile to achieve these two goals: increasing swing-phase peak torque while decreasing stance-phase peak torque. It does this by approximating human-exoskeleton interaction torques using angular velocity and angular acceleration (Eq. \eqref{eq_13}) when building the dynamics model.
Detailed results are shown in Fig. \ref{figa_torque_comparison}, which clearly illustrates that \texttt{ExoTraj} generates significantly higher peak torques during the swing phase. Future work can further study the mapping between exoskeleton assistive torques and real biological joint torques, and identify the optimal scaling strategy for ground-truth torques to minimize energy consumption.

\subsection{Results for Other Complex Scenarios}\label{other subject and sce}
This section presents supplementary analyses of the exoskeleton’s assistive performance across more complex task scenarios. Specifically, we report metabolic cost and heart rate measurements for movements including running and sit-to-stand transitions, and further evaluate the exoskeleton’s assistance efficacy for complex movements unseen in the training dataset.
Fig. \ref{figa7} compares metabolic rate and heart rate across three assistance modes during 10-degree ramp walking ($1.25$ m/s), running ($1.8$ m/s), and repeated sit-stand transition tasks.
The results confirm that the \texttt{ExoTraj} policy reduces both metrics across all scenarios:
1) Relative to Assist off, metabolic rate reductions range from $4.6\%$ (ramp walking) to $24.1\%$ (repeated sit-stand movements), while heart rate decreases by $8.7\%$ (running) to $16.1\%$ (ramp walking).
2) Compared with the No exo condition, metabolic rate and heart rate reductions span $0.4\%\thicksim11.1\%$ and $2.6\%\thicksim11.9\%$, respectively.
These findings further demonstrate that \texttt{ExoTraj} effectively lowers metabolic energy expenditure in human subjects, with a concurrent reduction in cardiovascular load.

\begin{figure*}
\centering
    \begin{minipage}{1\textwidth}
    \centering
    \includegraphics[width=1\textwidth]{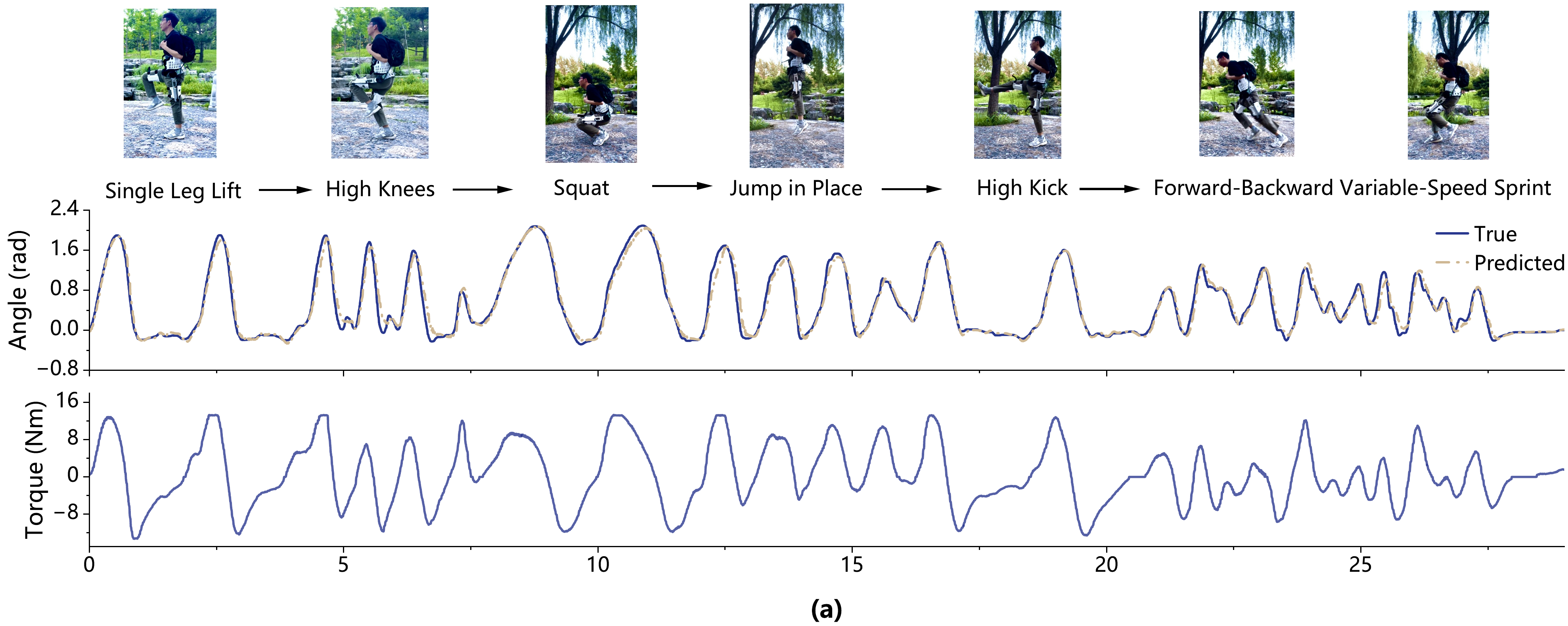}
  \end{minipage}
  \hfill
  \begin{minipage}{1\textwidth}
    \centering
    \includegraphics[width=1\textwidth]{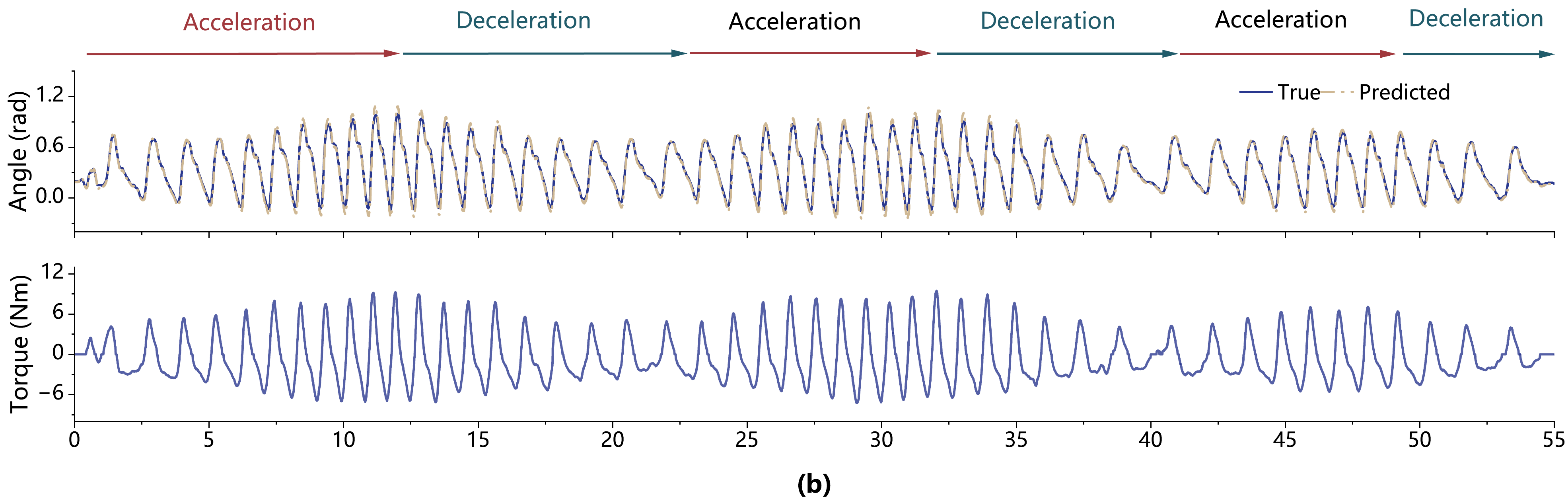}
  \end{minipage}    
    \captionsetup{justification=justified, singlelinecheck=false}
	\caption{\texttt{ExoTraj} torque assistance curve under complex scenarios and maneuvers. (a) shows the related curves for complex high-dynamic movements, including high knee lifts, rapid speed variations, jumping takeoffs, and single-leg hops, are absent from the training dataset. (b) shows the real-time prediction and torque assistance results under rapid speed variations on the treadmill.}. 
	\label{figa6}
\end{figure*}

Fig. \ref{figa6} presents torque profiles generated by the \texttt{ExoTraj} policy during exoskeleton assistance under varying treadmill speed conditions (b) and complex, high-dynamic movements (a). Fig. \ref{figa6} (b) shows that as locomotion speed increases, the peak magnitudes of the generated torque profiles increase and their frequency rises accordingly.  Meanwhile, the system demonstrates rapid adaptability to velocity variations: when the user comes to a complete stop, the assistance torque rapidly decays to zero with negligible latency. This demonstrates that policy \texttt{ExoTraj} is capable of adapting to dynamically changing tasks with rapid velocity variations. Fig. \ref{figa6} (a) presents the torque assistance curves of the \texttt{ExoTraj} policy for several motions unseen in the training dataset.
The experimental results consistently show that \texttt{ExoTraj} is capable of rapid adaptation to complex high-dynamic out-of-distribution tasks. The policy generates smooth torque and maintains precise real-time tracking performance even during high-velocity movements. These findings collectively validate the robust generalization ability of the control framework, making it well-suited for deployment in a wide range of dynamic outdoor operational environments.

\section{Other Discussion}\label{other discuss}
\subsection{Discussion for Generative methods}
Generative methods achieve precise generation of the target distribution by establishing a probabilistic path from an initial Gaussian distribution to the target paths. These methods can efficiently fuse data from heterogeneous distributions and enable highly generalizable generative prediction. The dominant paradigms in this field are diffusion models and flow matching.
Generative methods, such as diffusion policy (DP) \citep{chi2023diffusion,2025TROdiffusion}, which can excel at learning complex robotic tasks, have been extensively adopted for trajectory generation in robotic manipulation. However, DP suffers from two problems: slow inference speed and high training complexity. DP requires approximately $100$ denoising steps to generate a trajectory, and the inference time of DP typically requires approximately $1$ second \citep{ding2025}, which falls short of meeting the real-time control requirements of exoskeleton robots.  
Flow matching (FM) \citep{lipman2022flow} is a generative model method that leverages a learned vector field to map a simple prior distribution into a complex target distribution. From a high-level perspective, DP essentially involves solving a stochastic differential equation, whereas FM addresses an ordinary differential equation. Compared to DP, FM offers both easier training and faster inference while achieving comparable generation performance. 

However, if the FM generates precise trajectories, it still requires a considerable number of iterations (where the number of iterations in FM corresponds to the number of denoising steps in DP). An excessive number of iterations can compromise the real-time performance of the control system. To accelerate FM inference in image generation, \citet{geng2025mean} proposed using average velocity fields to reduce the number of sampling steps, and \citet{fransone} introduced a shortcut model to speed up inference. In the field of image generation, the objective is to produce images whose distribution is closely aligned with that of images within the same category, emphasizing similarity. In contrast, trajectory prediction prioritizes the precise and accurate generation of motion trajectories. Consequently, acceleration techniques developed for image generation are often inadequate for ensuring the accuracy of trajectory prediction under fast inference. Therefore, this paper designs the mechanism of trajectory feature incorporation to guide the learning direction of the FM, thereby enhancing the precision of trajectory generation with FM under fewer inference steps. 



\subsection{Discussion for Exoskeleton Assistance}
\subsubsection{Real-time Inference and Control:} Exoskeletons assist humans in complex environments with high dynamicity, imposing stringent requirements on inference and control, where the control frequency is typically no less than $200$ Hz. In perception and control pipelines, neural network inference is computationally expensive. For complex networks, the inference latency often exceeds $5$ ms, making the inference frequency lower than the required control frequency, which typically results in uncomfortable user experiences. Two approaches can address the aforementioned problem: first, model pruning and quantization, which inherently accelerate model inference speed; second, predictive triggering, which is only feasible for trajectory prediction (e.g., predicting $i$ steps and taking values starting from the $i$-th step of the trajectory when needed), but difficult to implement for single-step prediction. Therefore, when model parameters reach a certain scale, trajectory prediction becomes one of the important approaches to achieve real-time inference.

\subsubsection{Robust Control:}
When exoskeleton systems operate in outdoor environments, they are often subject to elevated levels of sensor measurement noise, which introduces localized errors and inconsistencies in the predicted motion trajectories. Directly feeding these noisy trajectories into the low-level controller as reference commands will inevitably result in control chattering, jerky movements, and degraded assistive performance. It is necessary to design a control optimization strategy based on the predicted trajectories to compensate for prediction noise and external noise disturbances, thereby achieving compliant and smooth control. Model predictive control is an optimization-based control method that operates on predefined objectives and relies on the human-exoskeleton coupled dynamic model. By incorporating carefully designed penalty terms into the optimization objective, we can significantly enhance the robustness of the generated torque against external disturbances. Although an exact human-exoskeleton coupled dynamic model cannot be derived, our experiments show that approximate dynamic parameters and interaction torque parameters can still be used to optimize torque curves that conform to human gait cycles.

\subsubsection{Cross-subject and Cross-task Prediction}:
Due to significant inter-subject variability in motor patterns and the complex and variable nature of exoskeleton assistance scenarios, the limited collected data cannot fully cover all possible motion distributions. Therefore, effectively improving the model's generalization prediction performance across subjects and tasks has become one of the core urgent problems in the field of intelligent exoskeleton assistance.
Currently, the development of foundation models has demonstrated strong generalization capabilities across tasks. Foundation models rely on large amounts of data and network parameters. Therefore, future research needs to collect more motion data from more subjects across more environmental tasks to enhance the generalization of exoskeleton trajectory prediction. Meanwhile, lightweight exoskeleton hardware is also required to reduce physical fatigue induced by long-duration data collection for human subjects.

\subsubsection{Personalized Assistance:}
When high-fidelity joint torque profiles are obtainable via motion capture systems, it remains an open question whether direct proportional scaling of these curves constitutes an optimal strategy for generating exoskeleton assistive torque profiles. In physiological joint torque patterns, the peak torque during the stance phase exceeds that of the swing phase, as the stance phase accommodates body weight support and postural balance. However, exoskeleton assistance is primarily designed to reduce user metabolic expenditure, and users derive greater metabolic benefit from external assistance during the swing phase of locomotion. Therefore, future investigations should explore a physiologically informed joint torque scaling paradigm to deliver personalized and metabolically efficient assistance.
When joint torques are not readily obtainable, a two-stage approach involving trajectory prediction and subsequent assistive torque generation has emerged as a promising solution for wearable exoskeleton control. Nevertheless, torques derived solely from kinematic trajectories may not align perfectly with individual users' neuromuscular characteristics. Optimizing torque profiles through iterative user preference feedback is a viable strategy to mitigate this limitation, but existing preference-based optimization frameworks depend on extensive user feedback data and exhibit prohibitively low sample efficiency. Accordingly, future research should prioritize the development of data-efficient torque preference optimization algorithms.

\end{document}